\documentclass{article}

\usepackage[preprint]{neurips_2025}
\usepackage{amsmath} 
\usepackage{graphicx} 
\usepackage{tabularx} 
\usepackage{subfigure}
\usepackage{colortbl}
\usepackage[table]{xcolor}
\usepackage{wrapfig}  

\usepackage[utf8]{inputenc} 
\usepackage[T1]{fontenc}    
\usepackage{hyperref}       
\usepackage{url}            
\usepackage{booktabs}       
\usepackage{amsfonts}       
\usepackage{nicefrac}       
\usepackage{microtype}      
\usepackage{xcolor}         

\newcommand{\systemname}{SpINRv2} 
\newcommand{\systemnamesansours}{SpINRv2 (Ours)}

\title{SpINRv2: Implicit Neural Representation for Passband FMCW Radars}

\author{%
  Harshvardhan Takawale\\
  Department of Computer Science\\
  University of Maryland, College Park \\
  \texttt{htakawal@umd.edu} \\
  \And
  Nirupam Roy \\
  Department of Computer Science \\
  University of Maryland, College Park \\
  \texttt{niruroy@umd.edu} \\
}

\begin{document}

\maketitle


\begin{abstract}
We present \textbf{\systemname{}}, a neural framework for high-fidelity volumetric reconstruction using Frequency-Modulated Continuous-Wave (FMCW) radar. Extending our prior work (\emph{SpINR}), this version introduces enhancements that allow accurate learning under high start frequencies—where phase aliasing and sub-bin ambiguity become prominent. Our core contribution is a fully differentiable frequency-domain forward model that captures the complex radar response using closed-form synthesis, paired with an implicit neural representation (INR) for continuous volumetric scene modeling. Unlike time-domain baselines, \systemname{} directly supervises the complex frequency spectrum, preserving spectral fidelity while drastically reducing computational overhead. Additionally, we introduce sparsity and smoothness regularization to disambiguate sub-bin ambiguities that arise at fine range resolutions. Experimental results show that \systemname{} significantly outperforms both classical and learning-based baselines, especially under high-frequency regimes, establishing a new benchmark for neural radar-based 3D imaging.
\end{abstract}


\section{Introduction}

Radar sensing has emerged as a resilient and cost-effective modality for 3D scene understanding, particularly in adverse environments where optical systems falter—such as fog, occlusion, or low light. Among radar types, \emph{FMCW (Frequency-Modulated Continuous-Wave)} radars stand out for their fine-grained range and velocity estimation using compact hardware. This has led to their widespread adoption in autonomous vehicles, robotics, and smart sensing platforms.

Despite this promise, conventional radar imaging pipelines—such as \emph{coherent backprojection}, range-Doppler processing, and voxelized SAR—suffer from critical limitations. They typically assume ideal signal propagation, ignore spectral effects like leakage and phase aliasing, and rely on dense aperture sampling and fixed voxel grids, making them inefficient or inaccurate under realistic constraints.

To overcome these challenges, we propose \textbf{\systemname{}}, an advancement over our original SpINR framework. Our approach marries \emph{physically-grounded signal modeling} with the expressive power of \emph{implicit neural representations (INRs)}. The key insight is to operate \emph{directly in the frequency domain}, where range is linearly encoded in beat frequency, enabling a \emph{closed-form differentiable forward model}. This allows us to synthesize only the frequency components relevant to the scene’s spatial extent—unlike time-domain models that simulate redundant data.

Crucially, \systemname{} improves upon its predecessor in two ways:
\begin{itemize}
    \item It extends to higher carrier frequencies, where shorter wavelengths introduce sub-bin ambiguity due to phase wrapping and resolution mismatch. We explicitly model these effects through \emph{complex-valued spectral synthesis} and mitigate them using \emph{smoothness and sparsity regularization}.
    \item It supports \emph{passband modeling}, capturing both amplitude and phase interactions crucial for generalization under realistic radar conditions.
\end{itemize}

Overall, \systemname{} offers:
\begin{itemize}
    \item \textbf{Spectral Supervision via Differentiable Forward Model} – avoids time-domain instability and models spectral leakage analytically.
    \item \textbf{Sub-voxel Continuous Representation} – learns scatterer fields without voxelization, enabling precise geometry recovery.
    \item \textbf{Scalability and Efficiency} – forward modeling is optimized for sparse apertures, computing only relevant frequency bins for supervision.
\end{itemize}

We validate \systemname{} through extensive experiments and ablations, showing that it outperforms classical backprojection, range-quantization models, and time-domain baselines across geometric and perceptual metrics. Our work sets a new precedent for neural radar reconstruction, especially under high-frequency regimes where prior methods degrade significantly.

\begin{figure}[t]
  \centering


  \includegraphics[width=\linewidth]{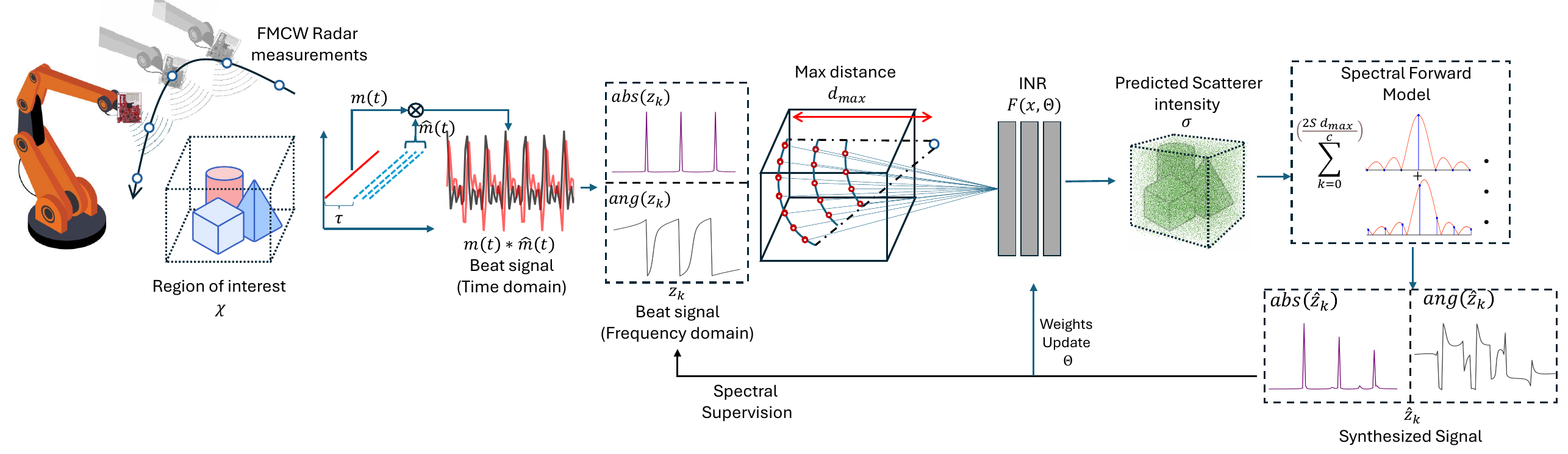}

  \caption{Overview of our method. FMCW radar measurements are collected from multiple viewpoints and transformed to the frequency domain for supervision. An implicit neural representation (INR) predicts scatterer intensity over the scene, and our differentiable spectral forward model synthesizes the complex-valued frequency-domain signal (magnitude and phase). The INR is trained by minimizing the discrepancy between synthesized and measured frequency-domain signals, using only the bins corresponding to the valid scene distances.}

  \label{fig:overview}
\end{figure}


\section{Related Work}

\subsection{FMCW Radar Imaging and Signal Modeling}

FMCW radar systems offer a lightweight, energy-efficient modality for range and velocity estimation, particularly well-suited for embedded applications in robotics and autonomous driving. The radar’s \emph{beat frequency encodes distance}, making it amenable to spectral methods like FFT for range estimation. Classical approaches such as backprojection~\cite{duersch2013backprojection}, range-Doppler processing~\cite{wagner2013wide}, and SAR~\cite{6504845} assume dense aperture sampling and ideal propagation, making them less robust to real-world imperfections. Recent learning-based efforts in radar imaging have largely focused on \emph{time-domain models} and coarse voxelized reconstructions~\cite{xu2022learned, sonny2024dynamic}. However, these neglect key spectral properties like \emph{bin mismatch}, \emph{phase spillage}, and \emph{DFT leakage}, which are essential for high-precision recovery.

\systemname{} differs fundamentally by building a \emph{native frequency-domain model}, capturing radar physics in a closed-form formulation and directly optimizing over spectral observations.

\subsection{Learning-Based Radar Reconstruction}

Deep learning has recently been applied to radar-based depth estimation~\cite{9506550}, occupancy prediction~\cite{2405.13307}, and neural beamforming~\cite{al2022review}. However, many of these approaches still operate in the \emph{time domain} and ignore spectral structure, limiting their ability to scale to high-resolution volumetric reconstruction. Our approach goes further by combining \emph{complex spectral supervision} with \emph{INRs}, resulting in higher geometric and perceptual fidelity.

\subsection{Implicit Neural Representations}

INRs like NeRF~\cite{10.1145/3503250} represent 3D scenes as continuous functions, mapping coordinates to intensity or radiance. While these models have revolutionized image-based view synthesis, recent extensions have explored their use in medical imaging~\cite{2407.02744}, audio~\cite{10.1145/3592141}, and transient signals~\cite{malik2023transient}.

Our work brings this paradigm into the \emph{radar domain}, modeling radar backscatter in the frequency domain using a differentiable forward model. \textbf{\systemname{}} is the first to integrate such spectral-aware INR modeling for radar signal reconstruction.

\subsection{MIMO Radar and Synthetic Aperture Systems}

Multi-Input Multi-Output (MIMO) radar systems and synthetic aperture setups improve angular and spatial resolution by synthesizing dense virtual arrays. Prior work~\cite{9136646} processes MIMO data using matched filtering or voxelized accumulation, often under uniform sampling. In contrast, \textbf{\systemname{}} supports \emph{non-uniform cylindrical sampling} and works directly with multistatic measurements—leveraging both diversity and sparsity in a differentiable pipeline.

\subsection{Classical Backprojection}

Backprojection is widely used in radar and CT imaging due to its physical interpretability~\cite{duersch2013backprojection}, but it has critical limitations: (1) it relies on uniform spatial discretization, leading to aliasing under sparse sampling; and (2) it is non-differentiable, making it unsuitable for learning-based pipelines. Our frequency-domain forward model addresses these limitations by explicitly modeling spectral structure and enabling gradient-based optimization of a continuous volumetric field. This leads to more accurate, high-resolution reconstructions, particularly under sparse and irregular sampling.

\section{Primer and Challenges}
\label{sec:primer}
\subsection{From Pulse-based Ranging to FMCW Radar}
Early neural volumetric reconstruction work on coherent sensors like \cite{10.1145/3592141} treats the sensing modality as a pulse–echo time-of-flight system. The returned waveform is a time-shifted copy whose delay $\tau$ is (after dechirping) directly proportional to the range.  Consequently, a forward model can be written as a convolution in time, and supervision in \emph{time domain} suffices.

Commercial mmWave hardware wavefront, however, employs frequency-modulated continuous-wave (FMCW) chirps rather than discrete pulses.  The device mixes the received chirp with the transmit replica on-chip, outputting only the \emph{beat} signal $b(t)$—a rapidly oscillating tone whose instantaneous frequency encodes round-trip delay \cite{7870764}.  In other words, whereas pulse radars measure \emph{time delay} that maps directly to distance, FMCW radars measure \emph{frequency} that directly maps to distance.

\subsection{Primer: FMCW Chirp Signal }
We develop our system model based on the mmWave radar sensors that transmits a signal with linearly time varying frequency. Such a signal is known as the Frequency modulated continuous wave signal (FMCW) and for a transceiver at location $(x_t,y_t,z_t)$ the signal can be formulated as - 
\[
m(t) = e^{\left(j2\pi\left(f_0 t + 0.5 S t^2\right)\right)}, \quad 0 \leq t \leq T,
\]
where $f0$ is the start frequency of the chirp and $S$ is the slope defined with units $Hz/s$. If there is a scatterer present at $(x_0,y_0,z_0)$ and reflects the signal with intensity $\sigma$ the received signal is delayed due to the distance traveled and scaled due to the propagation loss and $\sigma$ -
\[
\hat m(t) = \sigma m(t-\tau)  = \sigma e^{\left(j2\pi\left(f_0 (t-\tau) + 0.5 S (t-\tau)^2\right)\right)}, \quad 0 \leq t \leq T,
\]

where $\tau$ is the round-trip delay. If the distance between $(x_t,y_t,z_t)$ and $(x_0,y_0,z_0)$ is $\sqrt{(x_t - x_0)^2 + (y_t - y_0)^2 + (z_t - z_0)^2} = d$ $\tau$ is defined as $2d/C$. The FMCW radar mixes the transmitted and the received signal to demodulate the signal to a lower band. This is known as dechirping. The created beat signal is defined as - 
\begin{equation*}
m(t)* \hat m(t) = e^{(j2\pi f_0\tau + St\tau - 0.5Kt^2)}
\end{equation*}

the $-0.5Kt^2$ term is known as the residual video phase (RVP) and is often ignored as it is negligible~\cite{6810197} which simplifies the beat signal equation to-

\begin{align*}
m(t) * \hat{m}(t) = b(t) &= e^{j 2\pi (f_0 + S t)\tau} \nonumber \\
                  &= e^{j 2\pi f_0 \tau} \cdot e^{j 2\pi (S \tau) t}
\end{align*}

\subsubsection{Range Estimation and Resolution for FMCW Radar}

In an FMCW radar system, each transmitted chirp sweeps linearly from a start frequency \( f_0 \) over a bandwidth \( B \) in time \( T_c \). When the reflected signal is mixed with the transmitted replica, the result is a beat signal whose frequency \( f_b \) encodes the round-trip delay \( \tau \) to the target. This beat frequency has a direct physical interpretation:
\[
f_b = S \tau = S \cdot \frac{2d}{c},
\]
where \( S = \frac{B}{T_c} \) is the chirp slope, \( d \) is the range to the scatterer, and \( c \) is the speed of light. This tells us that \emph{range maps linearly to beat frequency}, a property that underlies the process of range compression. Applying an FFT to the beat signal transforms time delays into spectral components. Each frequency bin corresponds to a discrete round-trip delay, and hence to a spherical shell at fixed range. This frequency-domain view provides a clean and efficient mechanism for range estimation.

\vspace{0.5em}
\textbf{Range resolution}---the radar's ability to distinguish between two nearby scatterers in depth---is determined by the frequency resolution of the FFT $\Delta f = \frac{1}{T_c}$.
Using the beat frequency relationship, we can convert this to a resolution in space:
\[
\Delta f = \frac{2S \Delta d}{c} \quad \Rightarrow \quad \Delta d = \frac{c}{2S} \cdot \Delta f = \frac{c}{2B}.
\]
This expression reveals a useful insight: \emph{the range resolution depends only on the chirp bandwidth \( B \), not the chirp duration \( T_c \) or the start frequency \( f_0 \)}. Wider bandwidths yield finer depth resolution.

\subsection{Challenge 1: Learning from Time-Domain Signals is Fundamentally Ill-Conditioned}

\begin{wrapfigure}{r}{0.75\textwidth}
  \centering
  \begin{minipage}[b]{0.45\linewidth}
    \centering
    \includegraphics[width=\linewidth]{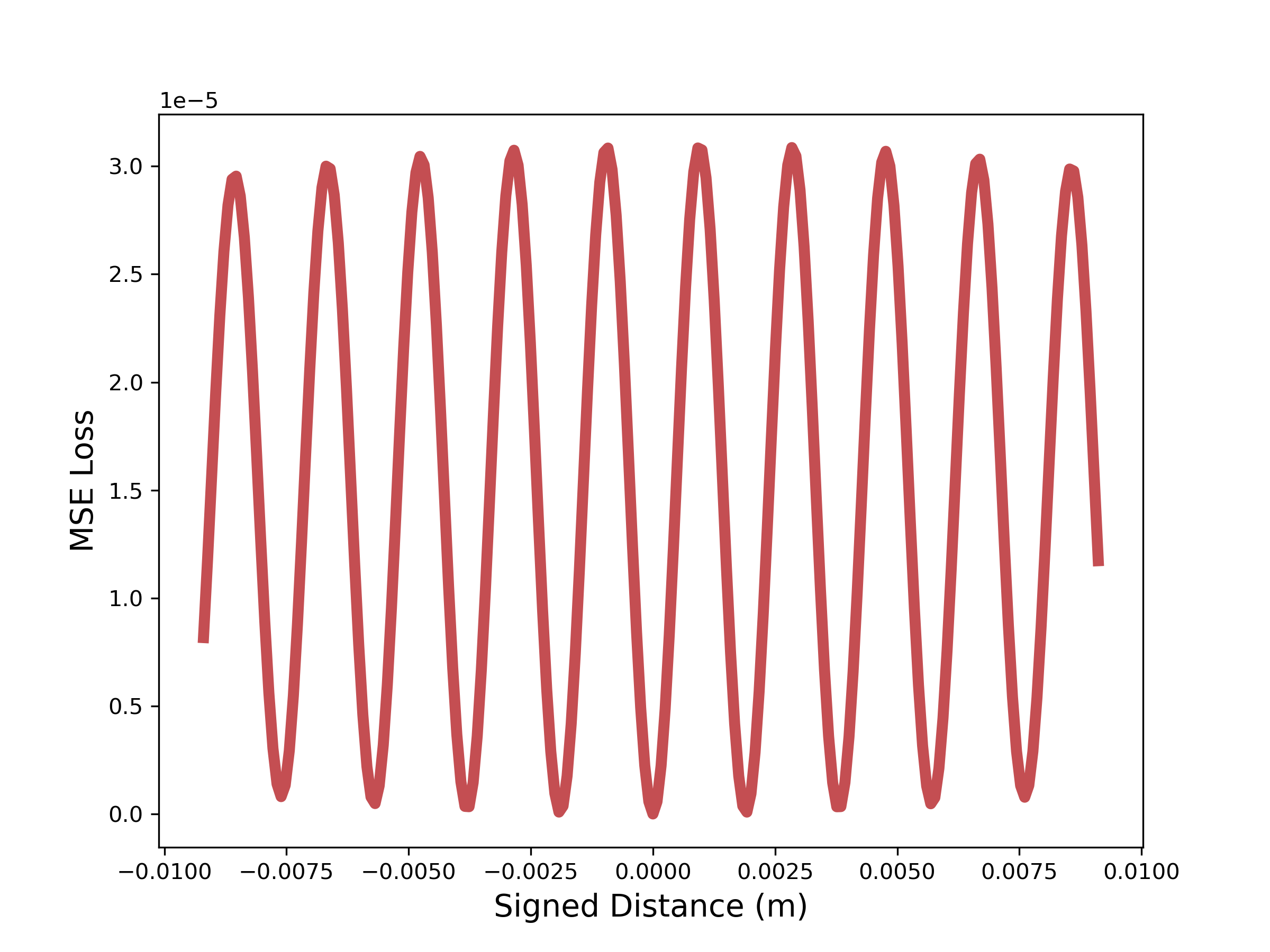}
  \end{minipage}
  \begin{minipage}[b]{0.45\linewidth}
    \centering
    \includegraphics[width=\linewidth]{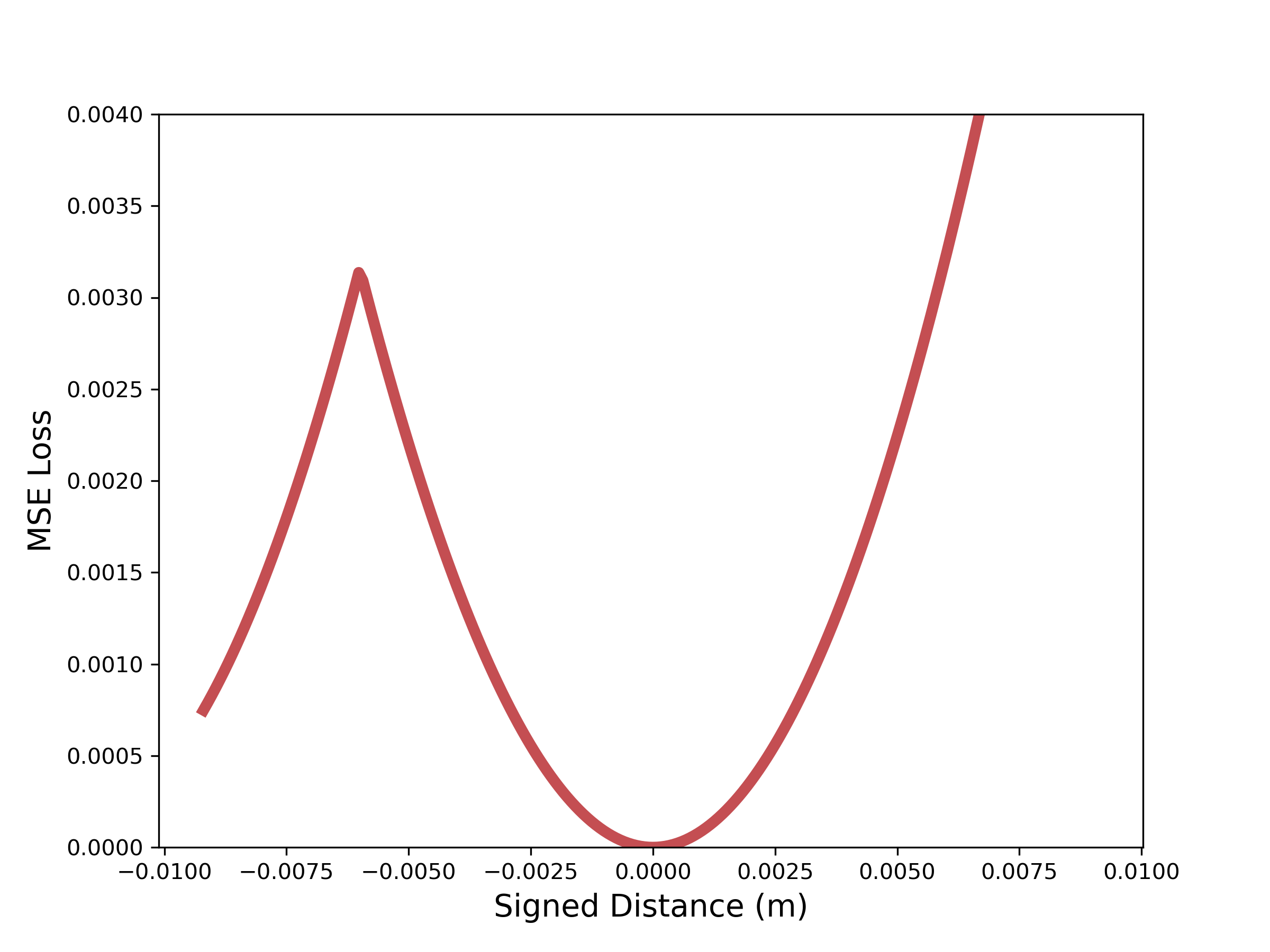}
  \end{minipage}
  \caption{Loss calculated on Time domain signal vs. loss calculated on Frequency domain signal. Time domain offers poor supervision.}
  \label{fig:td_vs_fd}
\end{wrapfigure}

A natural baseline for neural radar reconstruction is to follow the standard analysis-by-synthesis approach: simulate the full $N$-sample time-domain beat signal and supervise using an MSE loss. We refer to this approach as TF-TS (Time-domain Forward model with Temporal Supervision). This is also the signal used in classical backprojection method. However, the beat signal in the time domain is a rapidly oscillating sinusoid. Small shifts in scatterer range translate to phase shifts in the waveform—causing large and unpredictable changes in MSE. 

As a result, the loss landscape becomes highly non-convex, with sharp valleys and plateaus that lead to vanishing or exploding gradients. This instability makes optimization difficult and prevents the model from learning a meaningful reconstruction. Fig. \ref{fig:td_vs_fd} (left) illustrates how even small displacements can scramble the waveform, offering poor supervision. The fundamental problem lies in the mismatch between the nature of the time-domain signal and the loss function applied to it.

\subsection{Challenge 2: Frequency-Domain Alternatives Still Wasteful Without Proper Modeling}

Because range in FMCW radar maps linearly to beat frequency via $f_b = S\tau$, one intuitive improvement is to apply supervision in the frequency domain rather than the time domain. This leads to the TF-SS baseline (Time-domain Forward model with Spectral Supervision), which first simulates the full time-domain signal, then applies an FFT, and finally computes loss on the spectral bins. Spectral supervision is indeed more stable: a small shift in distance moves energy between frequency bins rather than scrambling all samples (Fig. \ref{fig:td_vs_fd}, right). However, TF-SS still simulates all $N=256$ samples and performs a full FFT—even though only a small number $K\ll N$ of frequency bins fall within the valid round-trip delay range of the scene. This leads to significant computational and memory overhead. A simpler but flawed alternative is to directly assign a scatterer's contribution to a frequency bin based on its range—what we term range-quantized supervision. However, this approximation neglects two critical signal properties: (i) the complex-valued nature of FMCW signals, where both phase and amplitude must be modeled, and (ii) spectral leakage, where energy smears into neighboring bins unless the delay perfectly aligns with a bin center.
\vspace{-5pt}
\paragraph{Summary of Challenges.}\mbox{}\\
\vspace{-1pt}
(1) \textbf{Time-domain instability.} The beat signal’s sinusoidal nature causes highly non-convex losses that are difficult to optimize. \\
(2) \textbf{Spectral inefficiency.} Conventional frequency-domain supervision (TF-SS) incurs unnecessary overhead by simulating full signals and performing large FFTs, even when most bins are unused. \\
(3) \textbf{Approximation pitfalls.} Naive range-quantized models fail to account for the complex-valued nature of signals and the spectral leakage induced by FFT discretization.

\systemname{} addresses these issues by working directly in the frequency domain using a closed-form forward model. This enables precise and efficient supervision on only the relevant spectral components, balancing physical accuracy with computational efficiency.

\section{Formulation of \systemname}

\subsection{Spectral Transformation}
\label{DFT}

In practical radar systems, signals are sampled and finite in duration. As such, spectral analysis is performed using the Discrete Fourier Transform (DFT), which approximates the Discrete-Time Fourier Transform (DTFT) at uniformly spaced frequency bins $\beta_k = \frac{2\pi k}{N}$.

To understand how DFT distributes energy across bins, consider a complex exponential signal $S_t = M e^{i(\alpha t + \phi)}$ with arbitrary angular frequency $\alpha$. The DFT evaluates to:
\[
Z_k = \frac{M}{N} e^{i\phi} \sum_{t=0}^{N-1} e^{i(\alpha - \beta_k)t}
= \frac{M}{N} e^{i\phi} \cdot \frac{1 - e^{i(\alpha - \beta_k)N}}{1 - e^{i(\alpha - \beta_k)}}
\]

This formulation can be rewritten in closed form by noting that $\beta_k N = 2\pi k$:
\begin{equation}
\boxed{
Z_k = \frac{M}{N} e^{i\phi} \cdot \frac{1 - e^{i \alpha N}}{1 - e^{i(\alpha - \beta_k)}}
}
\label{eq:zk_expression}
\end{equation}

This expression illustrates how energy at frequency $\alpha$ leaks into neighboring bins unless it perfectly aligns with a DFT bin center $\beta_k$. The extent of this leakage is explored next.

\subsubsection{Spectral Leakage and Energy Spillage}

The spectral transform discretizes the frequency axis into a set of equally spaced bins. When a scatterer lies at a distance such that its corresponding beat frequency aligns exactly with the center of a bin, all its energy is concentrated in that bin. This results in a clean, sharp spectral response.

However, in most practical cases, the beat frequency does not perfectly match a bin center. Instead, it falls somewhere between two bins. In such cases, the signal's energy does not remain localized but spreads across multiple neighboring bins. This phenomenon, known as \textit{spectral leakage}, is governed by the Dirichlet kernel:
\[
|Z_k| \propto \left| \frac{\sin\left(\frac{N}{2}(\alpha - \beta_k)\right)}{\sin\left(\frac{1}{2}(\alpha - \beta_k)\right)} \right|.
\]
The result is a sinc-like envelope centered at the true frequency, with energy distributed into adjacent bins. This introduces both amplitude and phase spillage across the spectrum. Importantly, this effect must be explicitly modeled in the forward process. Simplified approaches, such as range quantization, ignore this leakage and instead assign all energy to the nearest bin. Such approximations fail to capture the true complex domain spectral behavior of the signal and degrade reconstruction accuracy. This spectral leakage is a direct consequence of using finite-length DFTs and becomes especially critical in radar signal processing, where accurate frequency estimation directly impacts range resolution and reconstruction fidelity.

\subsection{Spectral Synthesis}
\label{sec:forward_model}

In this section, we present the frequency-domain forward model that serves as the basis for our implicit neural representation framework for volumetric reconstruction. Instead of modeling the time-domain signal and applying a DFT afterward, we directly model the complex frequency-domain response at each DFT bin, leveraging the closed-form DFT derivation introduced in Section~\ref{DFT}. This formulation is fully differentiable and thus amenable to gradient-based optimization used in neural rendering pipelines. Moreover, it encapsulates physical wave propagation, including transmission and backscattering, within the radar's operating bandwidth.

Let $\mathbf{x} \in \mathbb{R}^3$ denote a 3D point in the scene, and let $\sigma(\mathbf{x}) \in \mathbb{R}$ represent the reflectivity or scattering strength at that point. Define:
- $\mathcal{X} \subset \mathbb{R}^3$: the bounded spatial domain of interest,
- $\mathbf{o}_T, \mathbf{o}_R \in \mathbb{R}^3$: positions of the transmitter and receiver,
- $R_T(\mathbf{x}) = \|\mathbf{o}_T - \mathbf{x}\|$: propagation distance from Tx to $\mathbf{x}$,
- $R_R(\mathbf{x}) = \|\mathbf{o}_R - \mathbf{x}\|$: propagation distance from $\mathbf{x}$ to Rx,
- $\tau(\mathbf{x}) = (R_T(\mathbf{x}) + R_R(\mathbf{x}))/c$: total round-trip delay.

Then, the complex response at DFT bin index $k$ is given by:
\[
\boxed{
Z_k = \int_{\mathcal{X}} \frac{\sigma(\mathbf{x})}{N R_T(\mathbf{x}) R_R(\mathbf{x})} \cdot e^{i \phi(\mathbf{x})} \cdot \frac{1 - e^{i 2\pi S \tau(\mathbf{x}) N}}{1 - e^{i (2\pi S \tau(\mathbf{x}) - \beta_k)}} \, d\mathbf{x}
}
\]
where:
- $\beta_k = \frac{2\pi k}{N}$ is the angular frequency of the $k$-th DFT bin,
- $\phi(\mathbf{x}) = 2\pi f_0 \tau(\mathbf{x})$ is the baseband phase shift due to initial frequency $f_0$,
- $S$ is the chirp slope ($S = B/T_c$),
- $N$ is the number of time-domain samples (or DFT points). This forward model represents the coherent summation of returns from all scene points, modulated by the reflectivity $\sigma(\mathbf{x})$, and the frequency response derived from the DFT of a complex exponential tone with a delay $\tau(\mathbf{x})$. Given ground truth measurements $\tilde{Z}_k$, the loss is defined as:

\[
\mathcal{L} = \underbrace{\sum_k \left\| \left| Z_k \right| - \left| \tilde{Z}_k \right| \right\|_2^2}_{\mathcal{L}_{\text{mag}}} + \lambda \underbrace{\sum_k \left( \left\| \Re(Z_k) - \Re(\tilde{Z}_k) \right\|_2^2 + \left\| \Im(Z_k) - \Im(\tilde{Z}_k) \right\|_2^2 \right)}_{\mathcal{L}_{\text{phase}}}
\]

where $\lambda = 0.5$ controls the relative importance of magnitude vs. complex component supervision. 

This formulation can be extended to incorporate additional wave propagation phenomena, such as transmission attenuation models (e.g., based on material absorption), scattering probability as a function of incident angle or local geometry (e.g., Lambertian, specular, or volumetric models), and multipath interference or occlusion. These extensions can be embedded as additional multiplicative factors within the integrand.
\subsection{Smoothness and Sparsity Regularization}
\label{sec:regularization}

\begin{wrapfigure}{r}{0.75\textwidth}
  \centering
  \begin{minipage}[b]{0.45\linewidth}
    \centering
    \includegraphics[width=\linewidth]{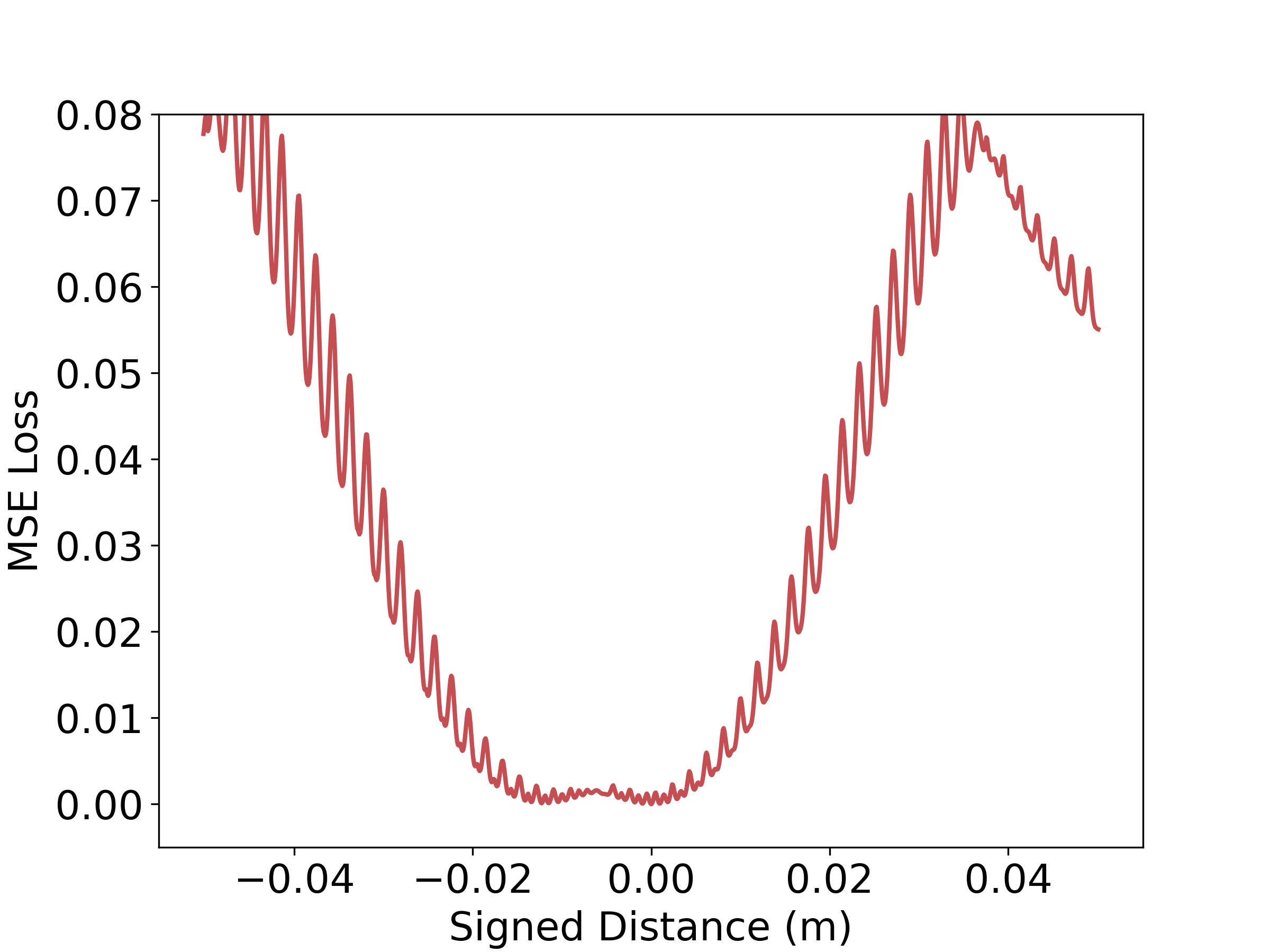}
  \end{minipage}
  \begin{minipage}[b]{0.45\linewidth}
    \centering
    \includegraphics[width=\linewidth]{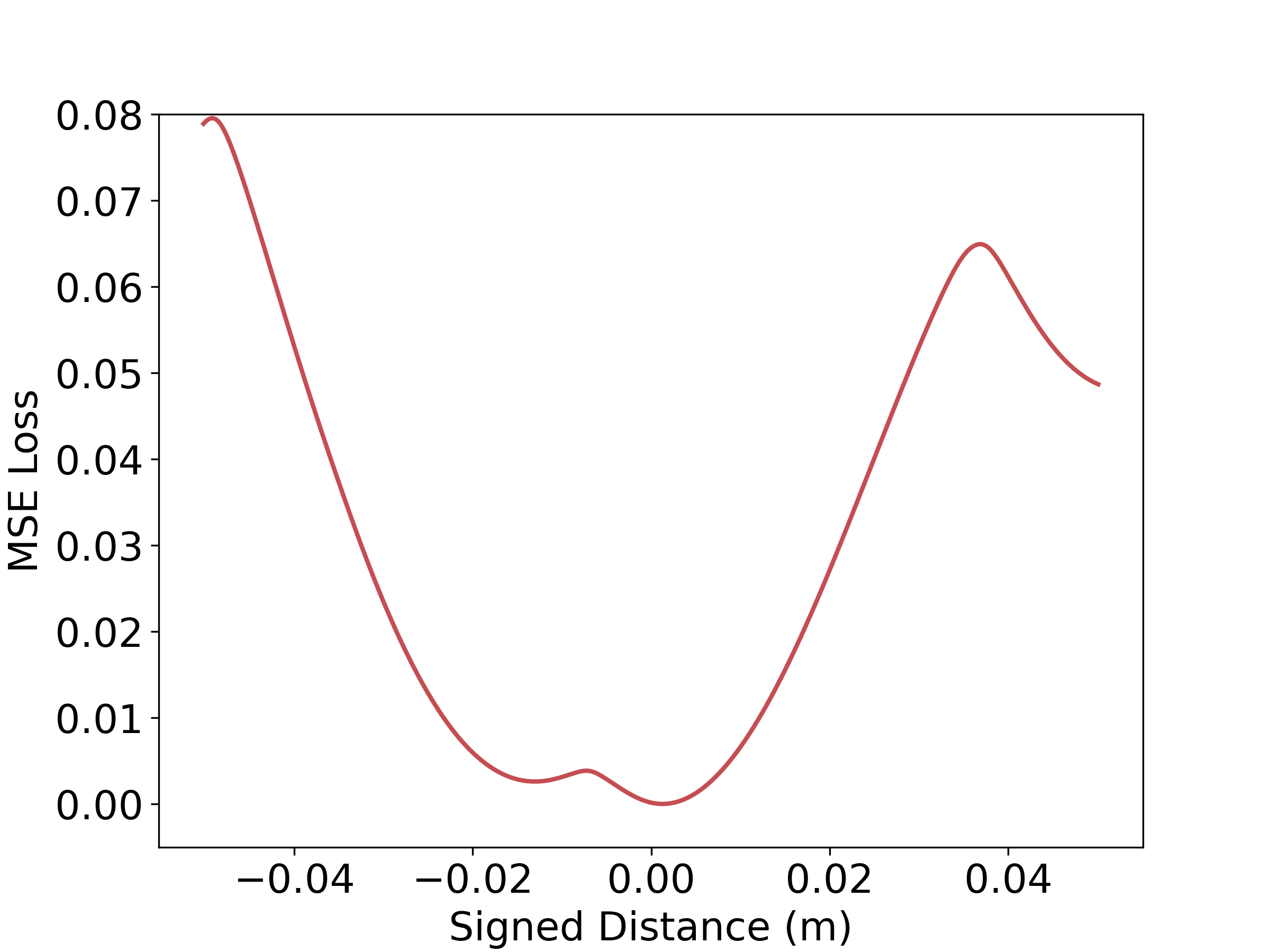}
  \end{minipage}
  \label{fig:loss_vs_sd_plots}
  \caption{Loss when \( \lambda/4 < \Delta r \)  vs. loss when \( \lambda/4 > \Delta r \). As the start frequency increases the loss landscaped becomes non-flat due to ambiguity.}
\end{wrapfigure}

While our frequency-domain forward model analytically captures the spectral response of a scene, it introduces a key challenge: ambiguity within each range bin. In FMCW radar, the center frequency of each FFT bin corresponds to a specific round-trip delay, and hence to a shell of possible scatterer locations. However, within a single bin---whose resolution is given by \( \Delta r = \tfrac{c}{2B} \)---there may exist multiple spatial configurations that produce similar frequency responses. This ambiguity is particularly pronounced when the radar wavelength \( \lambda \) is much smaller than the bin width, i.e., \( \lambda/4 < \Delta r \). In such cases, multiple sub-bin locations can yield nearly identical phase responses, resulting in a non flat loss landscape and unstable optimization.


Moreover, the start frequency \( f_0 \) of the chirp introduces an additional phase term \( 2\pi f_0 \tau \), which is independent of bin index. This global phase offset can further increase the number of scene configurations that produce the same spectral magnitude, especially in high-frequency regimes. These phase aliasing effects exacerbate the challenge of learning unique and spatially consistent reconstructions. To address these issues, we introduce two priors:

\textbf{(1) Smoothness Regularization.}
We encourage local spatial continuity in the predicted scattering field \( \sigma(\mathbf{x}) \) by penalizing its variation under small perturbations:
\[
\mathcal{L}_{\text{smooth}} = \mathbb{E}_{\delta \mathbf{x}} \left\| \sigma(\mathbf{x}) - \sigma(\mathbf{x} + \delta \mathbf{x}) \right\|_1,
\]
where \( \delta \mathbf{x} \sim \mathcal{U}(-\epsilon, \epsilon)^3 \) models small spatial offsets. This promotes coherent reconstructions and discourages sharp discontinuities.

\textbf{(2) Sparsity Regularization.}
Since most of the scanned volume is empty, we enforce sparsity in the scattering field by minimizing its \( \ell_1 \)-norm:
\[
\mathcal{L}_{\text{sparsity}} = \mathbb{E}_{\mathbf{x}} \left| \sigma(\mathbf{x}) \right|.
\]
This term biases the model toward compact reconstructions and suppresses ghost scatterers. Our final loss becomes:
\[
\mathcal{L}_{\text{total}} = \mathcal{L}_{\text{spectral}} + \beta \mathcal{L}_{\text{smooth}} + \gamma \mathcal{L}_{\text{sparsity}},
\]
where \( \beta \) and \( \gamma \) are tunable weights. These regularizers help disambiguate sub-bin solutions, improve convergence, and produce spatially plausible volumetric reconstructions, particularly under high-frequency radar sensing conditions.

\section{Experimental Setup}
\label{sec:experiment_setup}

We evaluate our frequency-domain forward model using simulated FMCW radar data over a cylindrical synthetic aperture. The setup mimics a practical configuration where the object rests on a turntable and the radar sensor is mounted on a vertical actuator. Their combined motion produces a dense 3D sampling of the scene from multiple viewpoints, forming a cylindrical inverse synthetic aperture. Our radar simulation follows the commertial TI AWR1843BOOST MIMO configuration, with a $3.585$~GHz bandwidth, $70.295 \times 10^{12}$~Hz/s chirp slope, and a sampling rate of $5$~MHz, yielding $256$ ADC samples per chirp. These time-domain beat signals are transformed via DFT into 256 frequency bins as discussed in Section~\ref{DFT}. The synthetic scene spans a $0.36$~m cube, with the radar positioned $0.23$~m from the object center. To obtain a well-structured dataset for learning, we first simulate $691{,}200$ multistatic MIMO radar measurements, then apply a multistatic-to-monostatic transformation to normalize Tx-Rx geometries and simulate a consistent forward model view. This sampling strategy closely mirrors real-world volumetric setups such as AirSAS~\cite{cowen2021airsas}, and provides a feasible path toward hardware realization compared to more idealized setups like spherical apertures. We train our models on RTXA5000 and iterate over all trans-receiver locations.

\section{Evaluation}
\label{sec:evaluation}

We evaluate our frequency-domain forward model for volumetric reconstruction by comparing it with multiple baseline approaches that vary in terms of signal representation (time vs. frequency domain) and physical modeling fidelity. All methods are implemented in our volumetric neural reconstruction pipeline using the same cylindrical scanning geometry described in Section~\ref{sec:experiment_setup}.We comparea \systemname{} with following baseline methods --

\begin{enumerate}
\item  Time-domain forward model with Temporal Supervision (TF-TS):
This baseline implements a physically accurate time-domain forward model based on path delays. The loss is computed between the simulated and predicted signals in the time domain, i.e., before applying any frequency transform. 
\item  Time-domain forward model with Spectral Supervision (TF-SS):
This method uses the same time-domain forward model as the baseline, but computes the loss after applying an FFT to convert the predicted signal into the frequency domain. This hybrid formulation partially accounts for spectral leakage and DFT artifacts.
\item  Range Quantization (RQ) model:
Inspired by traditional radar signal processing techniques, this method assigns each scatterer’s contribution to a DFT frequency bin based on its round-trip distance. The bin index is determined via quantized range thresholds, and scatterer contributions are summed into these bins to form a synthetic frequency-domain signal. While computationally efficient, this approximation ignores complex phasor interactions and introduces quantization artifacts. Loss is computed directly in the frequency domain.
\item  Coherent backprojection:
This traditional imaging method discretizes the 3D scene into voxels and estimates each voxel's intensity by coherently summing complex contributions from all Tx. Each contribution is phase-aligned based on the voxel's round-trip delay.
\end{enumerate}

\systemname{} directly models the radar response in the frequency domain using a differentiable, closed-form expression derived from the DFT of a delayed tone (Section~\ref{sec:forward_model}). This enables more accurate modeling of spectral leakage, non-bin-centered frequencies, and phase information, while being fully compatible with gradient-based optimization. The forward model is used in conjunction with a neural volumetric representation and optimized end-to-end.

\begin{figure}[t]
  \centering
  \setlength{\tabcolsep}{3pt}      
  \renewcommand{\arraystretch}{1}  

  \begin{tabular}{@{}c c c c c c c@{}}  
    \makebox[0.135\linewidth][c]{\small Range Quant.} &
    \makebox[0.135\linewidth][c]{\small TF-TS} &
    \makebox[0.135\linewidth][c]{\small TF-SS} &
    \makebox[0.135\linewidth][c]{\small Backproj.} &
    \makebox[0.135\linewidth][c]{\small \shortstack{SpINR\\(Baseband)}} &
    \makebox[0.135\linewidth][c]{\small \shortstack{\systemname\\(Passband)}} &    
    \makebox[0.135\linewidth][c]{\small GT} \\[2pt]

    \includegraphics[width=0.135\linewidth]{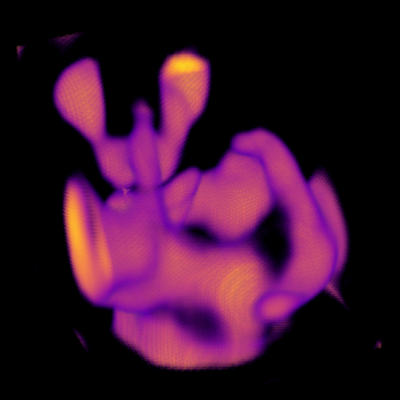} &
    \includegraphics[width=0.135\linewidth]{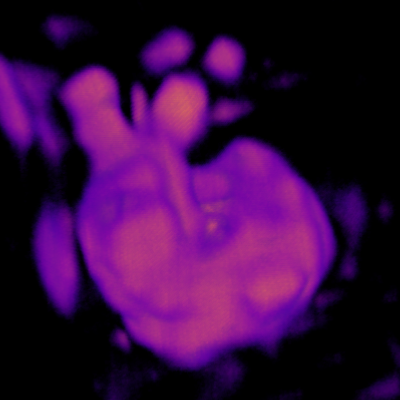} &
    \includegraphics[width=0.135\linewidth]{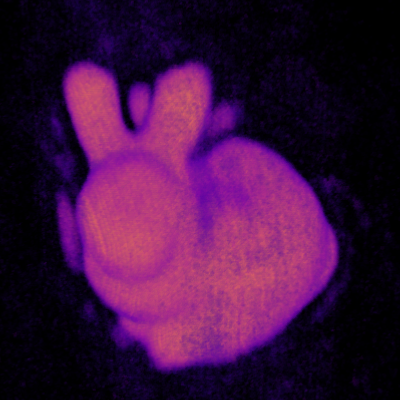} &
    \includegraphics[width=0.135\linewidth]{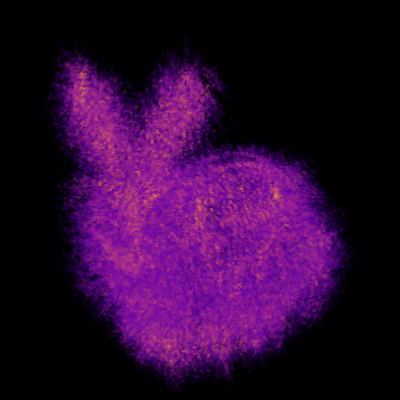} &
    \includegraphics[width=0.135\linewidth]{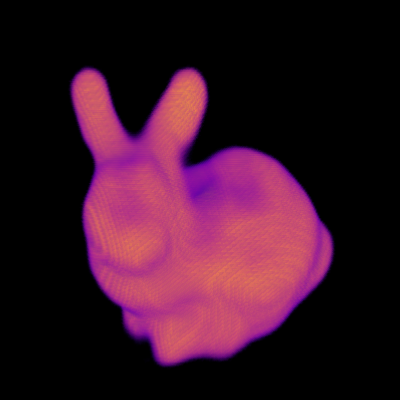} &
    \includegraphics[width=0.135\linewidth]{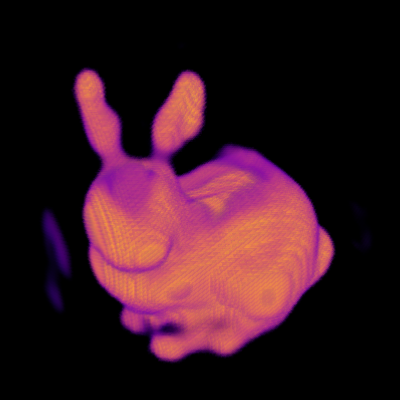} &    
    \includegraphics[width=0.135\linewidth]{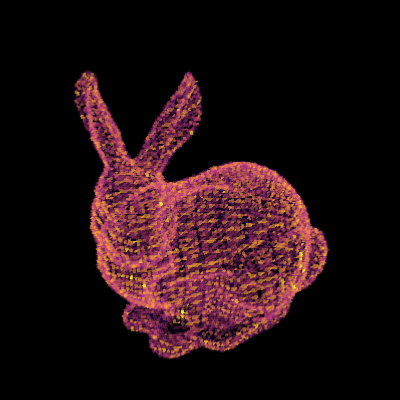} \\[3pt]

    \includegraphics[width=0.135\linewidth]{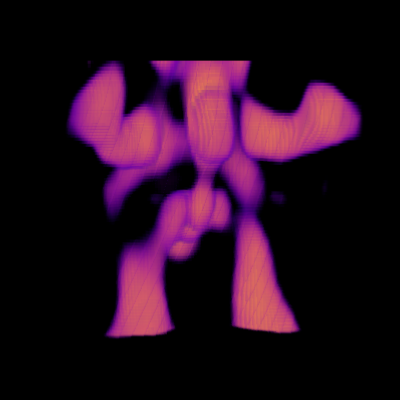} &
    \includegraphics[width=0.135\linewidth]{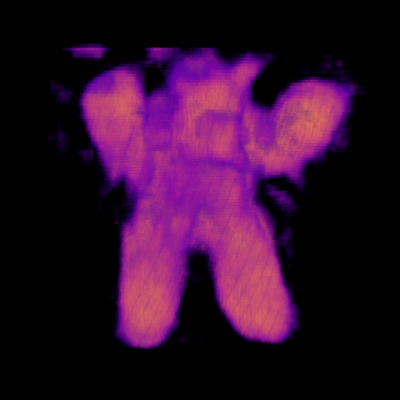} &
    \includegraphics[width=0.135\linewidth]{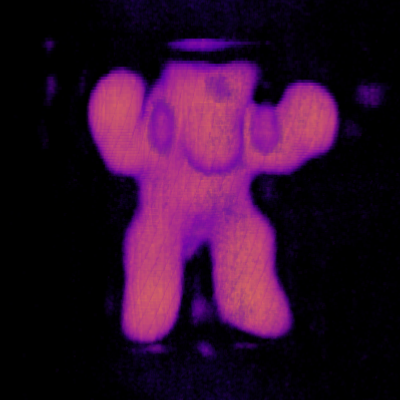} &
    \includegraphics[width=0.135\linewidth]{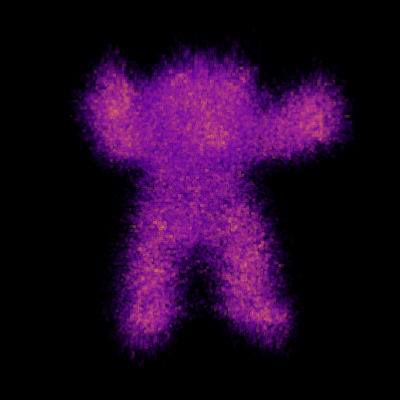} &
    \includegraphics[width=0.135\linewidth]{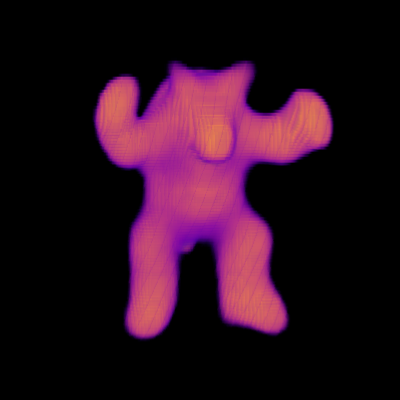} &
    \includegraphics[width=0.135\linewidth]{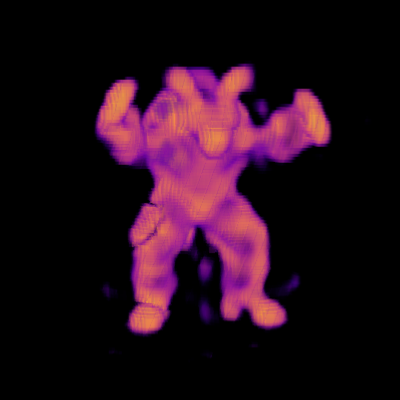} &    
    \includegraphics[width=0.135\linewidth]{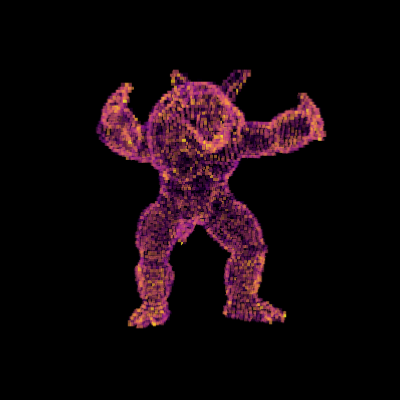} \\[3pt]

    \includegraphics[width=0.135\linewidth]{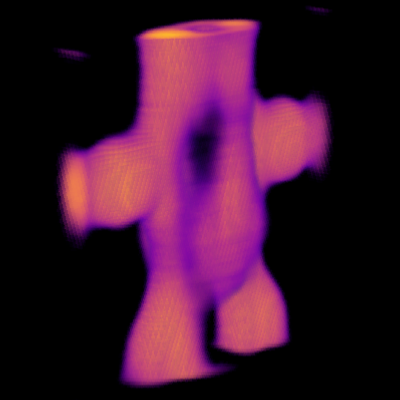} &
    \includegraphics[width=0.135\linewidth]{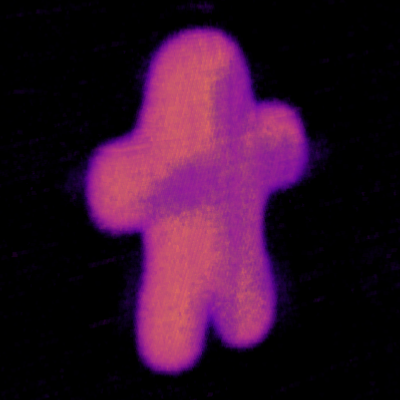} &
    \includegraphics[width=0.135\linewidth]{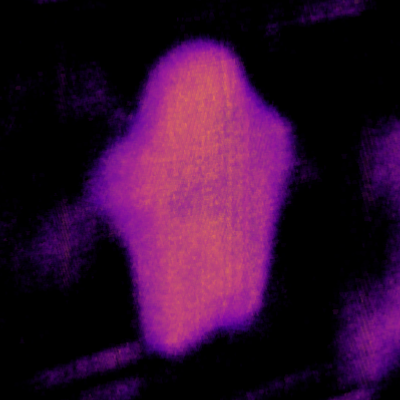} &
    \includegraphics[width=0.135\linewidth]{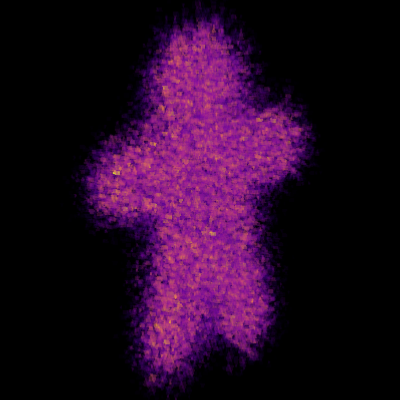} &
    \includegraphics[width=0.135\linewidth]{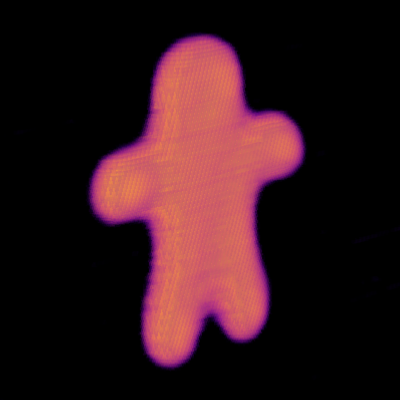} &
    \includegraphics[width=0.135\linewidth]{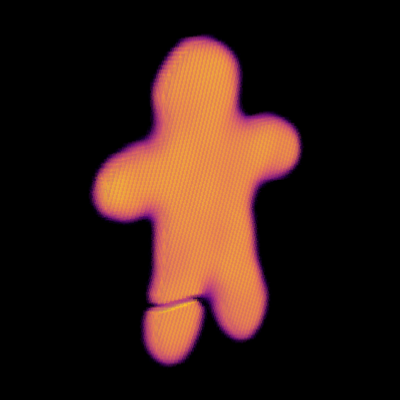} &    
    \includegraphics[width=0.135\linewidth]{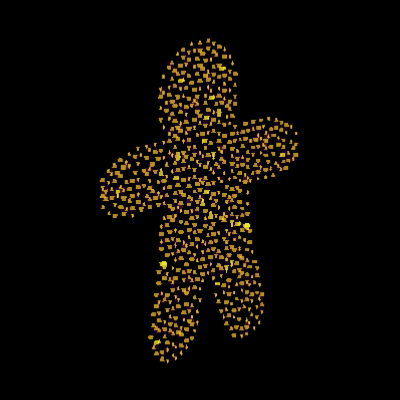} \\
  \end{tabular}

  \caption{Comparison of volumetric reconstructions for Range Quantization,
           Time-domain forward model with Temporal Supervision,
           Time-domain forward model with Spectral Supervision, and our method.
           Our reconstructions more accurately match the ground-truth geometry.}
  \label{fig:comparison}
\end{figure}

\subsection{Impact of Frequency-domain Supervision on Reconstruction}
Supervising reconstruction using the time-domain radar signal may seem natural, since that is what the hardware records. However, for FMCW radars, the beat signal is a rapidly oscillating sinusoid whose phase encodes the round-trip delay to each scatterer. This creates a significant problem: even a small change in scatterer location causes a large phase shift across the entire waveform.

From the perspective of a time-domain MSE loss, this shift makes the predicted signal appear drastically different from the ground truth—even though the underlying scene geometry is nearly unchanged. As a result, the loss landscape becomes highly non-convex, and gradients fail to provide a meaningful signal for optimization. The network struggles to learn a stable 1-to-1 mapping from scene geometry to signal, ultimately degrading reconstruction quality.

In contrast, the frequency domain provides a more robust supervision signal. Because beat frequency maps linearly to scatterer range, small geometric changes cause localized and interpretable changes in the spectrum, leading to smoother gradients and improved convergence. Figure~\ref{fig:comparison} illustrates the difference in reconstruction quality.


\begin{figure}[h]
  \centering
  \begin{minipage}[b]{0.5\linewidth}
    \centering
    \includegraphics[width=\linewidth]{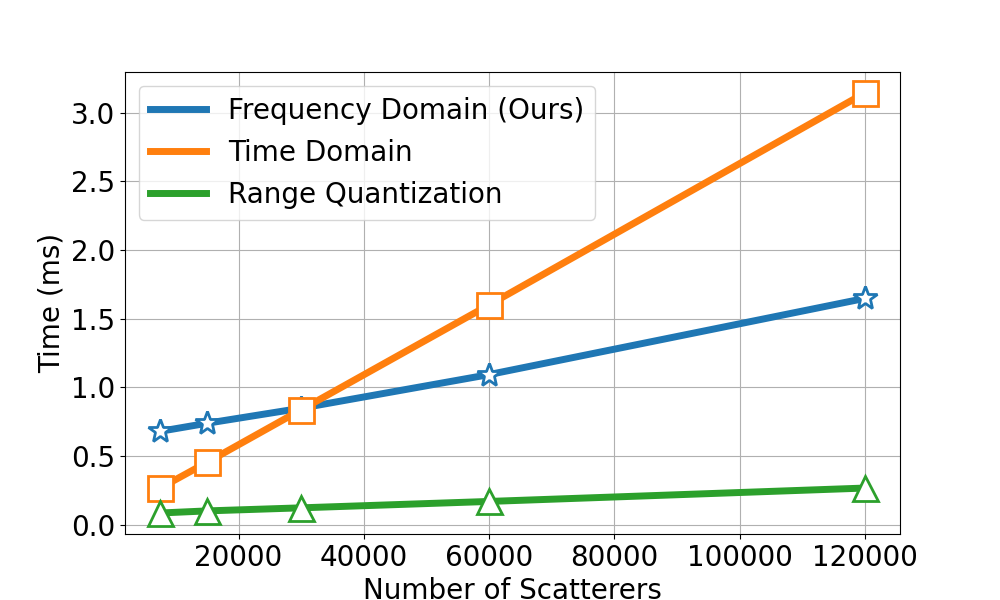}
    \caption{Comparison of runtime for different forward models. We measure the runtime with different scene sizes governed by the number of scatterers. Our proposed method outperforms the time domain forward model, while range quantization is the fastest.}
    \label{fig:forward_model_performancev2}
  \end{minipage}
  \hspace{0.04\linewidth}
  \begin{minipage}[b]{0.4\linewidth}
    \centering
    \includegraphics[width=\linewidth]{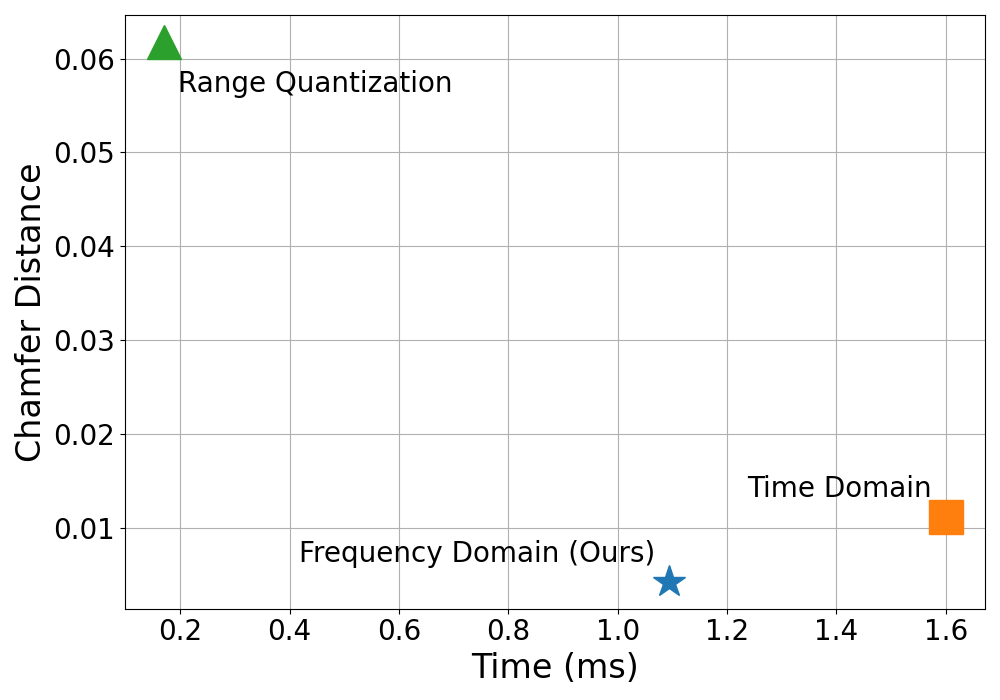}
    \caption{Comparison of the three forward models for their runtime and volume reconstruction accuracy. Even though the Range Quantization has the fastest runtime, it severely underperforms in volume reconstruction as signified by the chamfer distance (lower better).}
    \label{fig:Fdfwd_vs_Tdfwd2}
  \end{minipage}
\end{figure}

\subsection{Impact of Frequency-Domain Forward Model on Computation}
\label{sec:freq_vs_time}

In FMCW radar systems, range information is inherently encoded in the frequency domain: each scatterer's distance maps linearly to a beat frequency. When the spatial extent of the scene is bounded, this also bounds the set of relevant beat frequencies. In our setup, for example, only the first 16 frequency bins correspond to valid round-trip delays within the physical scene. This observation allows us to substantially simplify the forward model. In conventional time-domain formulations, the radar response is first synthesized as a full-length waveform—typically with 256 samples—and then transformed using an FFT to compute frequency-domain supervision losses. However, only a small fraction of these frequency bins are ultimately used for learning, making the majority of the computation redundant.

Our frequency-domain forward model addresses this inefficiency by directly computing the signal values at the required frequency bins using a closed-form expression derived from the Fourier transform of a delayed chirp. This formulation eliminates the need to simulate the full time-domain signal or apply an FFT, resulting in a leaner and more efficient synthesis process.As shown in Figure~\ref{fig:forward_model_performancev2}, this computational shortcut leads to a substantial reduction in forward-pass latency when evaluated on an NVIDIA RTX A5000 GPU. Importantly, this efficiency gain does not compromise accuracy; the frequency-domain formulation remains fully differentiable and retains the spectral fidelity necessary for high-quality volumetric reconstruction.

Together, these properties make the frequency-domain forward model not only physically motivated but also practically advantageous for scalable and real-time learning-based radar imaging.

\subsection{Impact of Frequency-Domain Forward Modeling on Learning Stability}
\label{sec:gradient_stability}

Although both our proposed frequency-domain forward model and the time-domain baseline ultimately compute loss in the frequency domain, we observe a significant performance gap in reconstruction accuracy and convergence speed. To investigate this, we analyze the gradient dynamics of the two approaches and show that our method yields smoother, more stable gradients throughout the network—enabling faster and more effective optimization.

\paragraph{Motivation from Prior Work.}
The importance of gradient flow in neural network design is well established. Deep architectures often suffer from vanishing or exploding gradients, motivating skip connections~\cite{10.5555/3305381.3305417}, normalized initialization~\cite{zhang2019fixup}, and sinusoidal activations for signal representation~\cite{10.5555/3495724.3496350}. These works demonstrate that preserving coherent and stable gradients across layers is crucial for trainability, especially in models with many layers or complex signal mappings. Similar reasoning has been used to justify architectural choices in implicit neural representations~\cite{10.5555/3495724.3496350}, and sparse pruning strategies~\cite{2002.07376}, Following this line, we examine how the formulation of the forward model affects gradient propagation in neural volumetric reconstruction.

\begin{figure}[th]
    \centering
    \subfigure[Mean]{
        \includegraphics[width=0.45\linewidth]{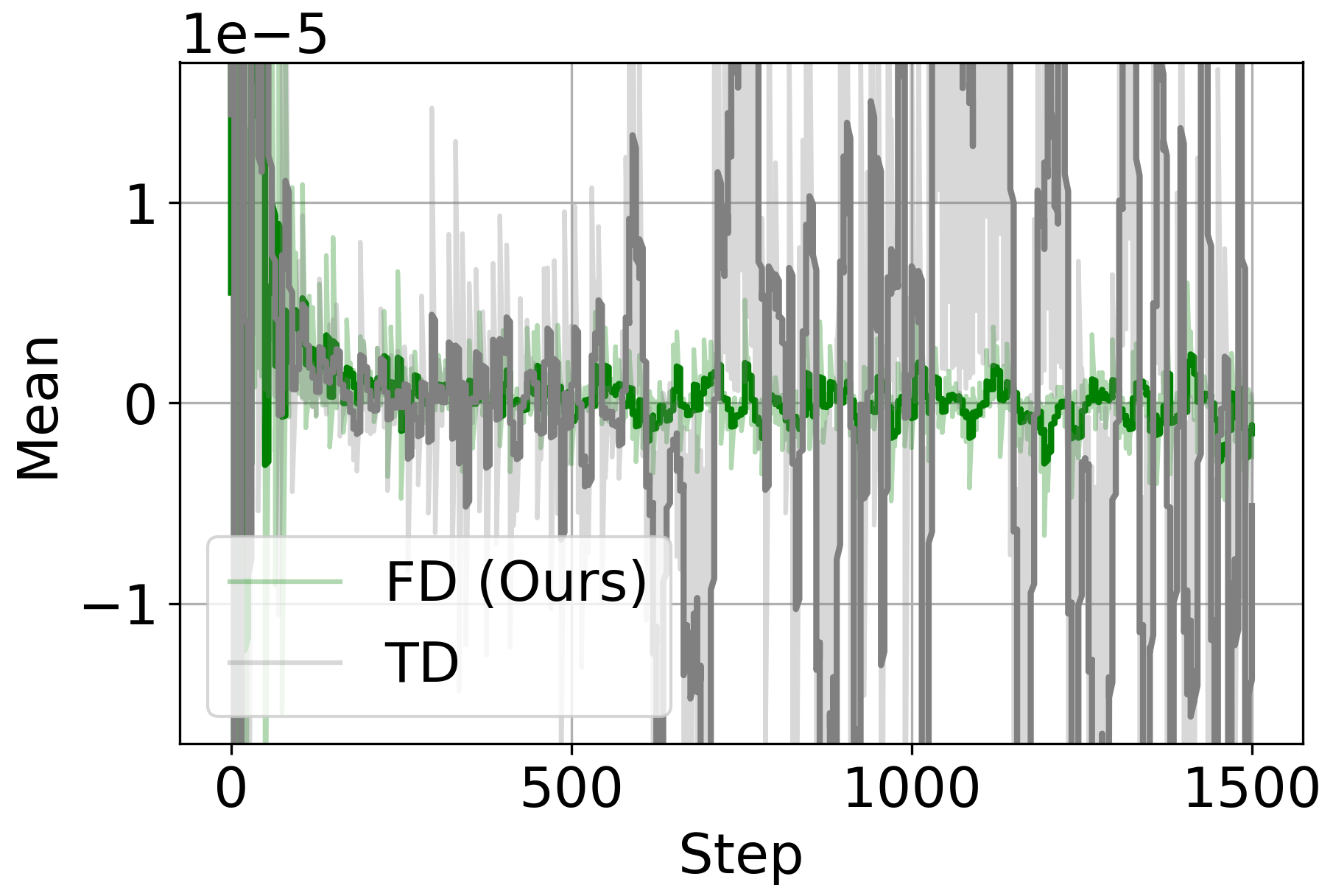}
    }%
    \subfigure[Standard Deviation]{
        \includegraphics[width=0.45\linewidth]{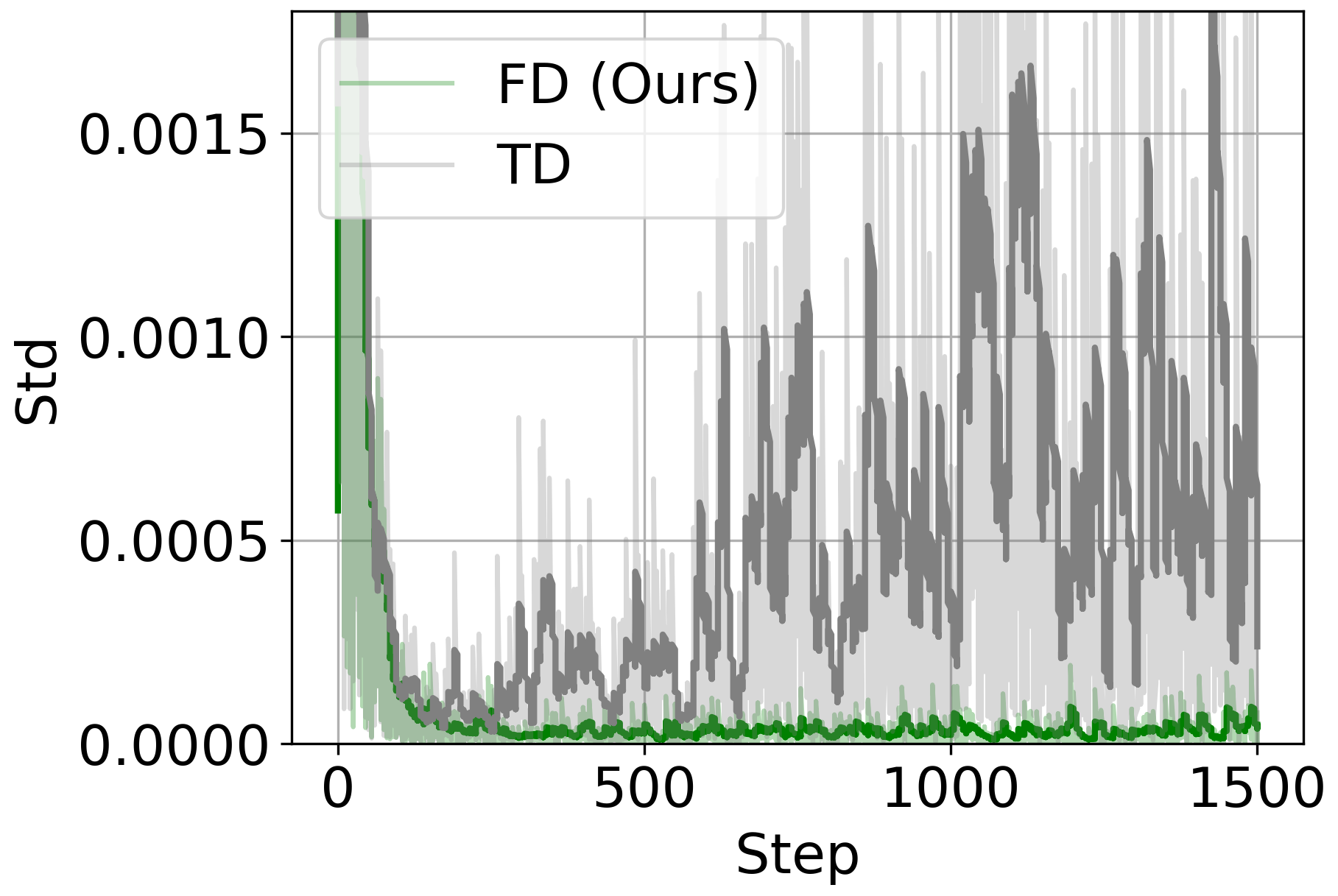}
    }\\
    \caption{Comparison between (a) Mean and (b) standard deviation of the gradients for the first layer wrt the number of training steps for Frequency domain reconstruction (FD) and Time domain reconstruction (TD). The mean and standard deviation are more stable for our proposed method. For the time domain forward model the the gradient tends to explode.  We observe a similar trend for all the subsequent layers.}
    \label{fig:gradient_experiment}
\end{figure}

We visualize the computation graph of both forward models (Figure~\ref{fig:gradient_experiment}) and observe that the time-domain model includes additional operations (e.g., time-domain simulation, full FFT) that introduce longer, more fragmented backpropagation paths. In contrast, our frequency-domain forward model analytically predicts only the relevant frequency bin values, resulting in a shallower and more direct gradient path to the implicit neural representation. To quantify this, we log the mean and standard deviation of the weight gradients at the first layer of the INR across training. Our method (green) exhibits significantly more stable gradient statistics than the time-domain model (gray). The variance remains controlled throughout training, avoiding sudden spikes or vanishing behavior. 

These findings mirror prior observations from studies on residual networks~\cite{10.5555/3305381.3305417}, gradient confusion~\cite{10.5555/3524938.3525723}, and signal representations~\cite{10.5555/3495724.3496350}, where architectural decisions that preserve or align gradients lead to better optimization and generalization. Our work extends this perspective to signal modeling for FMCW radar, showing that a forward model grounded in spectral domain physics not only improves interpretability but also facilitates learning through more stable and effective gradient flow.

\subsection{\systemname~ vs. Classical Coherent Backprojection}
\label{sec:vs_backprojection}

To further validate the effectiveness of our proposed frequency-domain forward model, we compare it against a classical reconstruction baseline: coherent backprojection (CBP). This method discretizes the scene into voxels and accumulates backprojected energy from all Tx paths by aligning phases based on path delays. While physically grounded, CBP has significant limitations that our learning-based approach addresses. \\
(1) Inherent Discretization and Aliasing:  
CBP relies on uniform voxel grids and discrete binning of round-trip delays, which often leads to aliasing artifacts and blurring, especially when the sampling aperture is non-uniform or sparse. Our method, by contrast, uses a continuous volumetric representation (via neural fields) and avoids explicit voxelization, enabling sub-voxel accuracy in both representation and rendering. \\
(2) Lack of Differentiability and Learning:  
CBP is a purely geometric method—it does not learn from data or optimize a forward model. As a result, it is highly sensitive to aperture coverage, noise, and partial views. Our method incorporates a fully differentiable, physics-informed forward model, enabling gradient-based optimization that refines the scene representation to best match all measurements. As shown in our quantitative results (Tables~\ref{tab:evaluation},~\ref{tab:chamfer_distance}), \systemname{} achieves significantly better reconstruction compared to CBP.






\textbf{Quantitative Results.} We evaluate the quality of 3D volumetric reconstruction across five methods using standard metrics widely adopted in computer vision and inverse rendering. These include Intersection-over-Union (IoU), Chamfer Distance (CD), Hausdorff Distance (HD), Peak Signal-to-Noise Ratio (PSNR), Structural Similarity Index (SSIM), and Learned Perceptual Image Patch Similarity (LPIPS). Together, they capture both geometric fidelity and perceptual image quality by comparing voxel grids, rendered views, and point cloud representations of reconstructed shapes.

We compare our method, \systemname{}, against: (i) a time-domain forward model with time-domain loss (TF-TS), (ii) a time-domain forward model supervised in the frequency domain (TF-SS), (iii) a reconstruction-by-querying baseline (RQ), and (iv) classical backprojection. All methods are trained on the same synthetic cylindrical radar dataset described in Section~\ref{sec:experiment_setup}.

Table~\ref{tab:evaluation} summarizes average performance across all test scenes. Our model achieves the best performance across all six metrics, showing clear improvements in both geometric reconstruction (IoU, CD, HD) and view-based similarity (PSNR, SSIM, LPIPS). Table~\ref{tab:chamfer_distance} presents Chamfer Distance per object, where \systemname{} outperforms all baselines with consistent gains across diverse shapes. Top 3 results in each column are highlighted with the color scheme - best in red, 2nd in orange, 3rd in yellow.


\definecolor{best}{rgb}{1.0, 0.8, 0.8}       
\definecolor{secondbest}{rgb}{1.0, 0.9, 0.7}  
\definecolor{thirdbest}{rgb}{1.0, 1.0, 0.6}   

\begin{table}[t]
\centering
\setlength{\tabcolsep}{5pt}
\renewcommand{\arraystretch}{1.2}
\begin{tabular}{lcccccc}
\toprule
\textbf{Method}       & IoU ($\uparrow$)           & Chamfer ($\downarrow$)        & Hausdorff ($\downarrow$)       & PSNR ($\uparrow$)          & SSIM ($\uparrow$)          & LPIPS ($\downarrow$)        \\
\midrule
\textbf{\systemnamesansours{}}& \cellcolor{best}0.0908          & \cellcolor{best}0.0055          & \cellcolor{secondbest}0.0713    & \cellcolor{best}17.06          & \cellcolor{best}0.801           & \cellcolor{best}0.248 \\
TF-TS                 & \cellcolor{thirdbest}0.0483                          & 0.0219     & 0.1470     & 9.27      & 0.469      & 0.683   \\
TF-SS & 0.0177 & \cellcolor{thirdbest}0.0100 & \cellcolor{thirdbest}0.0722 & \cellcolor{secondbest}13.5134 & \cellcolor{secondbest}0.7189 & \cellcolor{secondbest}0.3917 \\
RQ                    & 0.0135                          & 0.0728                          & 0.1856                          & 6.27                           & 0.390                          & 0.806                        \\
Backproj             & \cellcolor{secondbest}0.0598     & \cellcolor{secondbest}0.0099   & \cellcolor{best}0.0461           & \cellcolor{thirdbest}11.28     & \cellcolor{thirdbest}0.691     & \cellcolor{thirdbest}0.426\\
\bottomrule
\end{tabular}
\caption{Average reconstruction performance across all scenes.}
\label{tab:evaluation}
\end{table}


\vspace{-2mm}
\begin{table}[t]
\centering
\setlength{\tabcolsep}{4pt}
\renewcommand{\arraystretch}{1.2}
\begin{tabular}{lccccccc}
\toprule
\textbf{Method}       & bunny             & spot              & lucy              & armadillo         & dragon            & woody             & teapot            \\
\midrule
\textbf{\systemnamesansours{}} & \cellcolor{best}0.0050       & \cellcolor{best}0.0066       & \cellcolor{best}0.0044    & \cellcolor{best}0.0042    & \cellcolor{best}0.0042    & \cellcolor{best}0.0061    & \cellcolor{best}0.0080    \\
TF-TS         & 0.0111                       & 0.0594                       & \cellcolor{secondbest}0.0055 & 0.0196                    & \cellcolor{secondbest}0.0056 & \cellcolor{thirdbest}0.0110 & 0.0408                    \\
TF-SS & \cellcolor{secondbest}0.0067 & \cellcolor{thirdbest}0.0172 & \cellcolor{thirdbest}0.0062 & \cellcolor{secondbest}0.0055 & \cellcolor{thirdbest}0.0059 & 0.0128 & \cellcolor{thirdbest}0.0159 \\
RQ            & 0.0593                       & 0.0859                       & 0.0690                  & 0.0618                    & 0.0625                    & 0.0671                    & 0.1041                    \\
Backproj      & \cellcolor{thirdbest}0.0095  & \cellcolor{secondbest}0.0160 & 0.0074 & \cellcolor{thirdbest}0.0079 & 0.0072                  & \cellcolor{secondbest}0.0090 & \cellcolor{secondbest}0.0124 \\
\bottomrule
\end{tabular}
\caption{Chamfer Distance for each object (lower is better).}
\label{tab:chamfer_distance}
\end{table}





\subsection{Ablation Study}
\subsubsection{Effect of Smoothness and Sparsity Regularization}

In the high-frequency regime, where \( \lambda/4 < \tfrac{c}{2B} \), our model begins to exhibit reconstruction artifacts in the form of faint "shells" around the true geometry. These arise from the fact that multiple spatial locations can produce similar phase responses within a bin—particularly when only a single transmitter is considered.

\begin{wrapfigure}{r}{0.4\textwidth}
  \vspace{-12pt}
  \centering
  \setlength{\tabcolsep}{1pt}
  \begin{tabular}{cc}
    \multicolumn{2}{c}{\small Without Regularization} \\
    \includegraphics[width=0.18\textwidth]{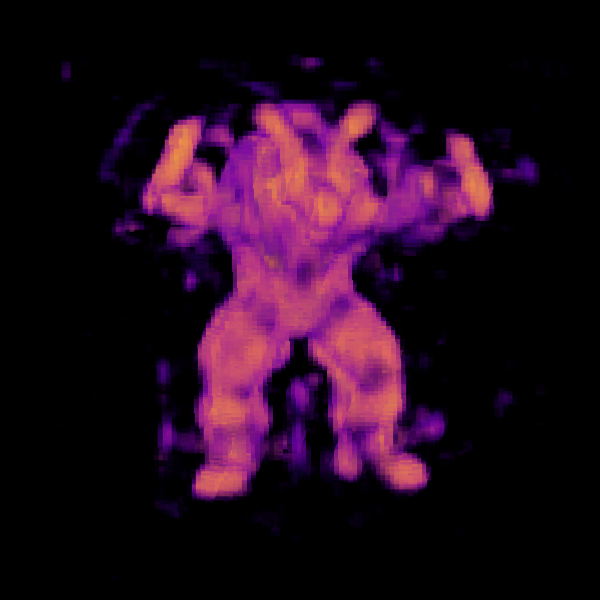} &
    \includegraphics[width=0.18\textwidth]{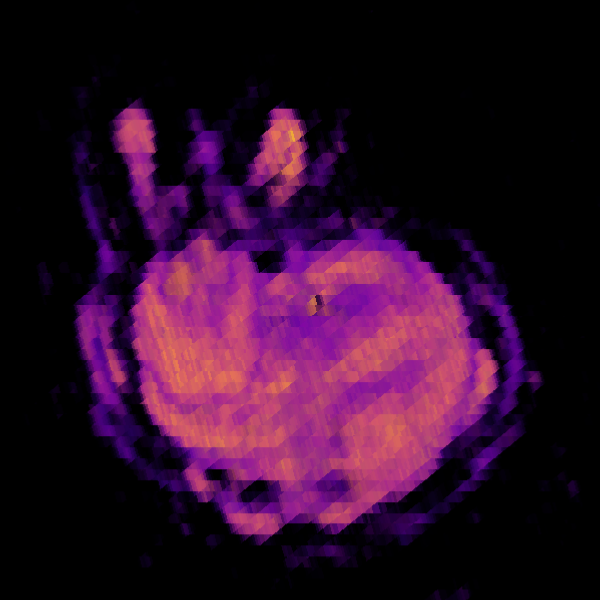} \\[-2pt]
    \includegraphics[width=0.18\textwidth]{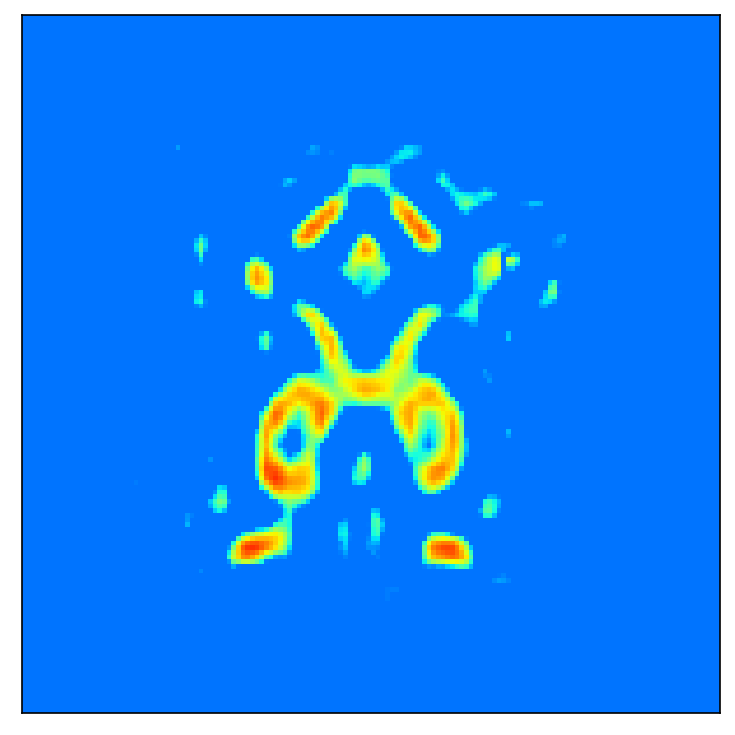} &
    \includegraphics[width=0.18\textwidth]{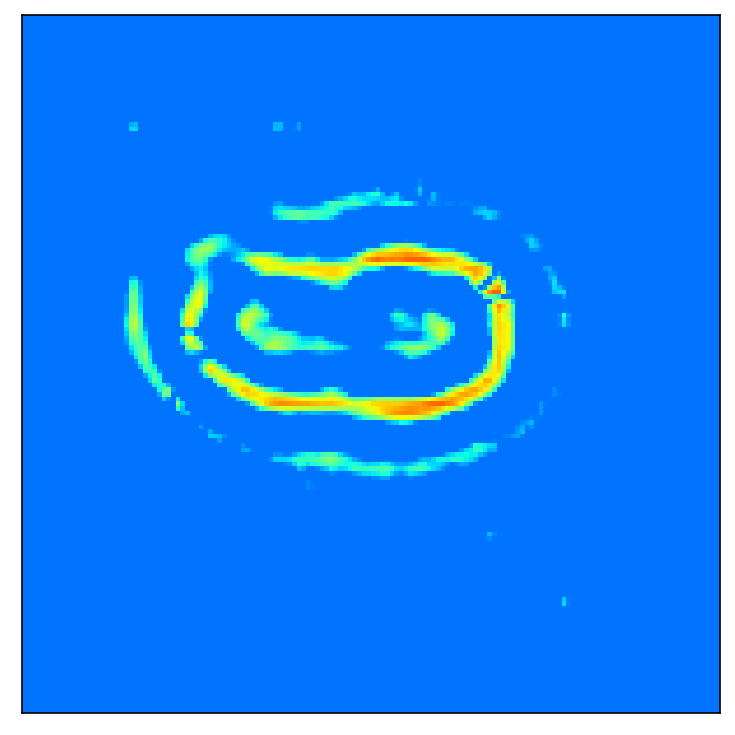} \\[2pt]
    \multicolumn{2}{c}{\small With Regularization} \\
    \includegraphics[width=0.18\textwidth]{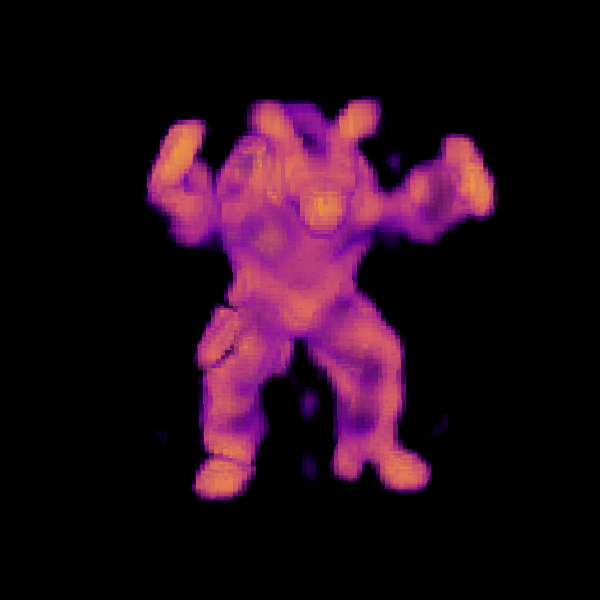} &
    \includegraphics[width=0.18\textwidth]{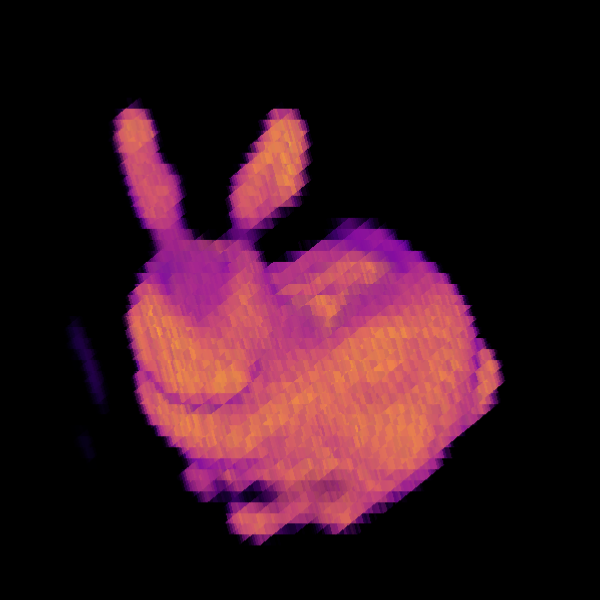} \\[-2pt]
    \includegraphics[width=0.18\textwidth]{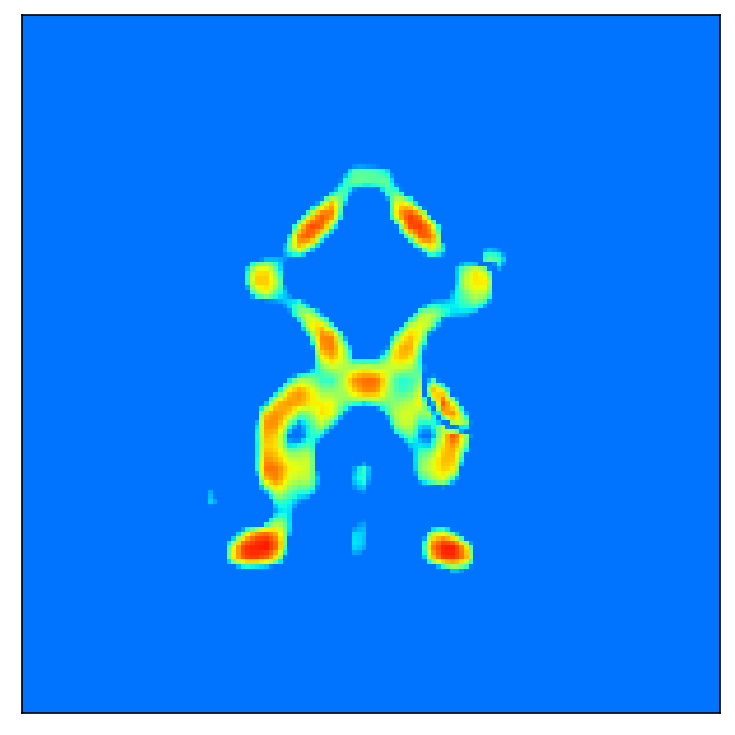} &
    \includegraphics[width=0.18\textwidth]{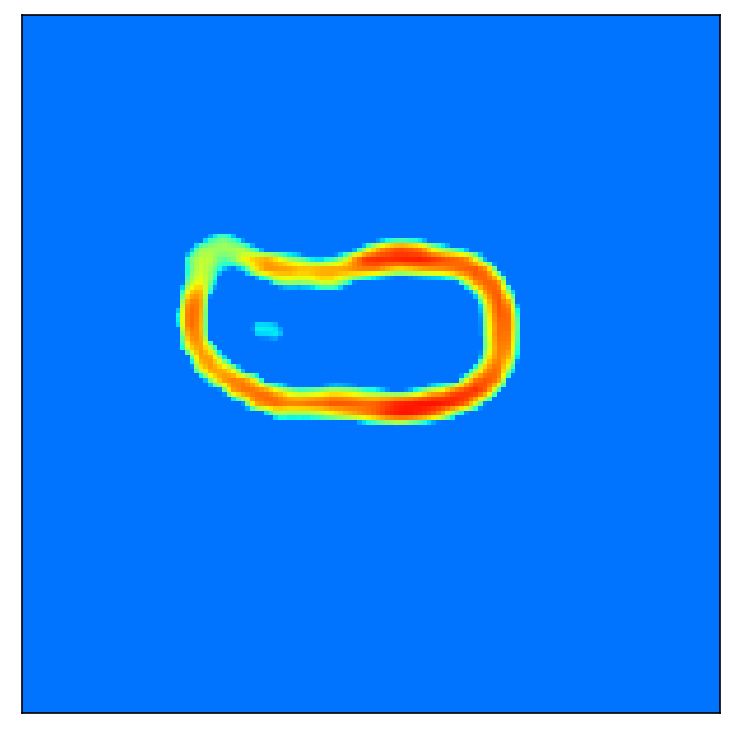} \\
  \end{tabular}
  \vspace{-5pt}
  \caption{Effect of smoothness and sparsity regularization on reconstruction quality. Top row shows the 3D reconstruction for various objects and the bottom row shows a 2D slice plot of the same object. Regularization reduces noise, eliminates shell artifacts, and improves surface coherence.}
  \label{fig:regularization_comparison}
  \vspace{-8pt}
\end{wrapfigure}

However, as this fast-moving phase is distance-based, even though an incorrect scatterer location's phase response matches for one or few transmitter positions, only the true surface satisfies the response across \emph{all} transmitter positions. As a result, these ambiguous shell responses tend to be weaker and more spatially inconsistent. This observation suggests a natural strategy for disambiguation: enforce global consistency and suppress incoherent energy.

To that end, we introduce smoothness and sparsity regularization. Smoothness encourages the predicted scatterer field to vary continuously in space, removing offset shells and fragmentary surfaces. Sparsity, on the other hand, biases the model toward compact volumetric support, suppressing low-energy artifacts and eliminating redundant ghost structures.

As shown in Fig.~\ref{fig:regularization_comparison}, reconstructions trained without regularization exhibit fragmented, noisy surfaces and disconnected shell-like structures. Adding regularization dramatically improves geometric fidelity—cleaning up the surface, sharpening edges, and removing spurious responses while preserving the fine details of the true object.

\subsubsection{Effect of Bandwidth Change on Reconstruction Quality}

The spatial resolution of FMCW radar systems is fundamentally limited by the chirp bandwidth \( B \), with the minimum distinguishable range given by \( \tfrac{c}{2B} \). As the bandwidth decreases, this resolution limit coarsens, which traditionally leads to severe blurring in reconstruction techniques such as backprojection. However, \systemname{} is not bound by explicit voxel grids or range quantization. Instead, it learns a continuous implicit representation of the scatterer field. This allows the network to infer and preserve spatial structure even under low-bandwidth conditions, where conventional methods lose detail.

\begin{wrapfigure}{r}{0.65\textwidth}
  \centering
  \setlength{\tabcolsep}{1pt}      
  \renewcommand{\arraystretch}{1}  

  \begin{tabular}{@{}c c c c c@{}}  
    & \makebox[0.2\linewidth][c]{\small 40 MHz} &
      \makebox[0.2\linewidth][c]{\small 400 MHz} &
      \makebox[0.2\linewidth][c]{\small 800 MHz} &
      \makebox[0.2\linewidth][c]{\small 4 GHz} \\[3pt]

    \makebox[0.08\linewidth][r]{\raisebox{25pt}{\small \textbf{\systemname{}}}} &
    \includegraphics[width=0.2\linewidth]{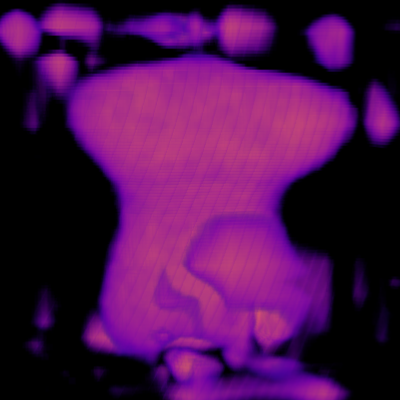} &
    \includegraphics[width=0.2\linewidth]{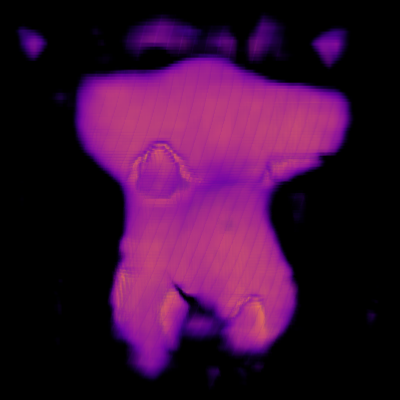} &
    \includegraphics[width=0.2\linewidth]{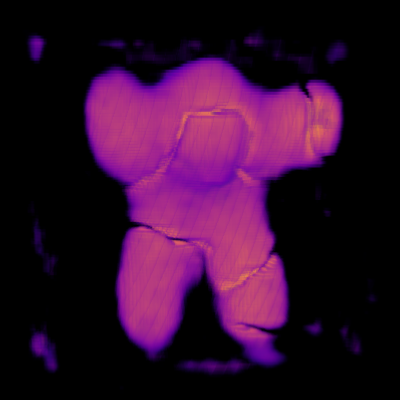} &
    \includegraphics[width=0.2\linewidth]{Figures/armadillo/volumes_comparison_armadillo_withbackprojection_4.png} \\[1pt]

    \makebox[0.08\linewidth][r]{\raisebox{25pt}{\small \textbf{Backproj.}}} &
    \includegraphics[width=0.2\linewidth]{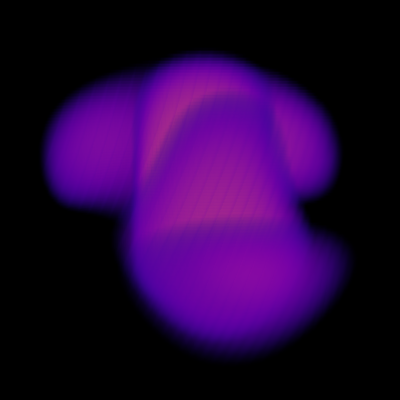} &
    \includegraphics[width=0.2\linewidth]{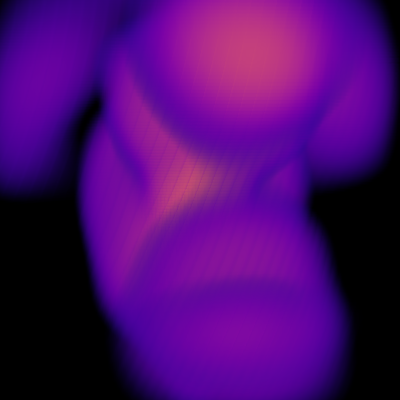} &
    \includegraphics[width=0.2\linewidth]{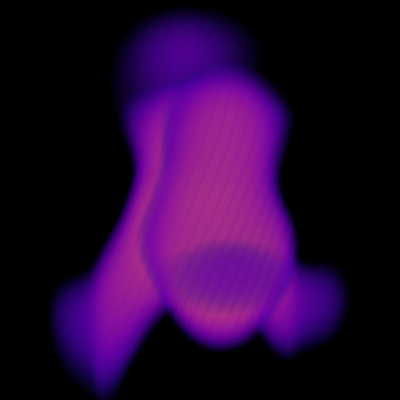} &
    \includegraphics[width=0.2\linewidth]{Figures/armadillo/volumes_comparison_armadillo_withbackprojection_3.png} \\
  \end{tabular}

  \caption{Comparison of volumetric reconstructions for different bandwidths (4GHz, 800 MHz, 400 MHz, 40 MHz) using \systemname{} and classical backprojection. \systemname{} shows a graceful degradation as compared with backprojection.}
  \label{fig:bandwidth_change_experiment}
\end{wrapfigure}

Figure~\ref{fig:bandwidth_change_experiment} compares reconstructions from \systemname{} and classical backprojection as the bandwidth is reduced from 4 GHz down to 40 MHz—a 100x decrease. While both methods suffer some degradation in spatial sharpness, \systemname{} exhibits a notably graceful degradation. Key object features remain preserved at lower bandwidths, and the reconstructions retain their overall shape and topology. In contrast, backprojection rapidly collapses into severely blurred and aliased outputs. These results highlight the robustness of the learned representation in \systemname{} to physical resolution limits imposed by the sensor, demonstrating its utility in bandwidth-constrained regimes.

\subsubsection{Effect of Start Frequency on Reconstruction Quality}
While the range resolution of an FMCW radar depends only on the chirp bandwidth \( B \), the starting frequency \( f_0 \) introduces an additional phase offset into the frequency-domain signal. As shown in our closed-form expression for the bin response \( Z_k \), this manifests as a multiplicative term \( e^{i\phi} \), where \( \phi = 2\pi f_0 \tau \). It becomes critical when learning to map spatial coordinates to bin responses, particularly due to the periodic nature of \( e^{i\phi} \).

This phase ambiguity becomes problematic when \(\lambda/4 \) falls below the bin resolution \( \tfrac{c}{2B} \). In this regime, multiple sub-bin scatterer positions can generate indistinguishable responses in the spectral domain, making it difficult for the model to identify a unique solution. The result is a degraded reconstruction with faint "shadow shells" appearing around the true object geometry.
\par

We empirically evaluate this effect by fixing the bandwidth \( B \) and varying \( f_0 \) from 1 GHz to 5 GHz. At lower start frequencies, where \( \lambda/4 > \tfrac{c}{2B} \), the model reconstructs clean and compact geometry. As \( f_0 \) increases and the wavelength decreases, the reconstructions progressively degrade, showing increasing artifacts, duplication, and spatial blur due to phase aliasing. This smooth transition highlights the critical role of start frequency in determining the identifiability and fidelity of volumetric radar reconstructions. In the next section we show how regularization helps with higher frequencies.
\begin{wrapfigure}{r}{0.65\textwidth}
  \vspace{-10pt}
  \centering
  \includegraphics[width=0.2\textwidth]{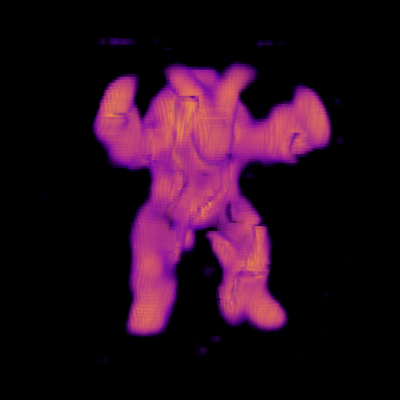}
  \includegraphics[width=0.2\textwidth]{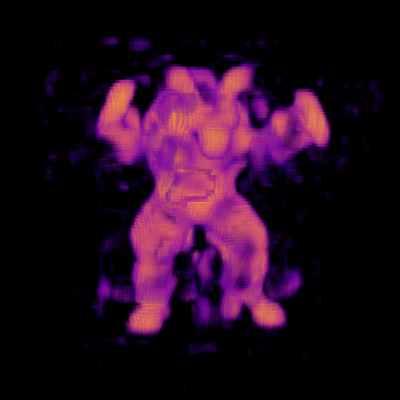}
  \includegraphics[width=0.2\textwidth]{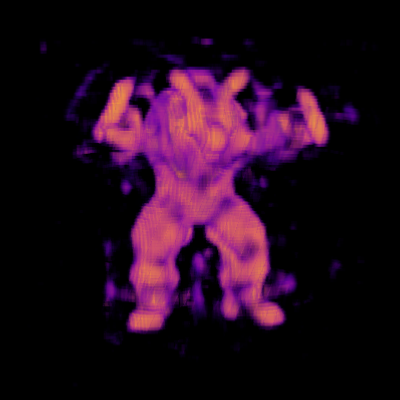}
  \caption{Effect of different start frequencies 2, 3, 4 GHz}
  \label{fig:startfreq_change_experiment}
  \vspace{-10pt}
\end{wrapfigure}

\subsubsection{Evaluation on a High-Frequency Sheet Benchmark}
\label{sec:coarse_fine_eval}

\begin{figure}[h]
  \centering
  \begin{minipage}[b]{0.23\linewidth}
    \centering
    \includegraphics[width=\linewidth]{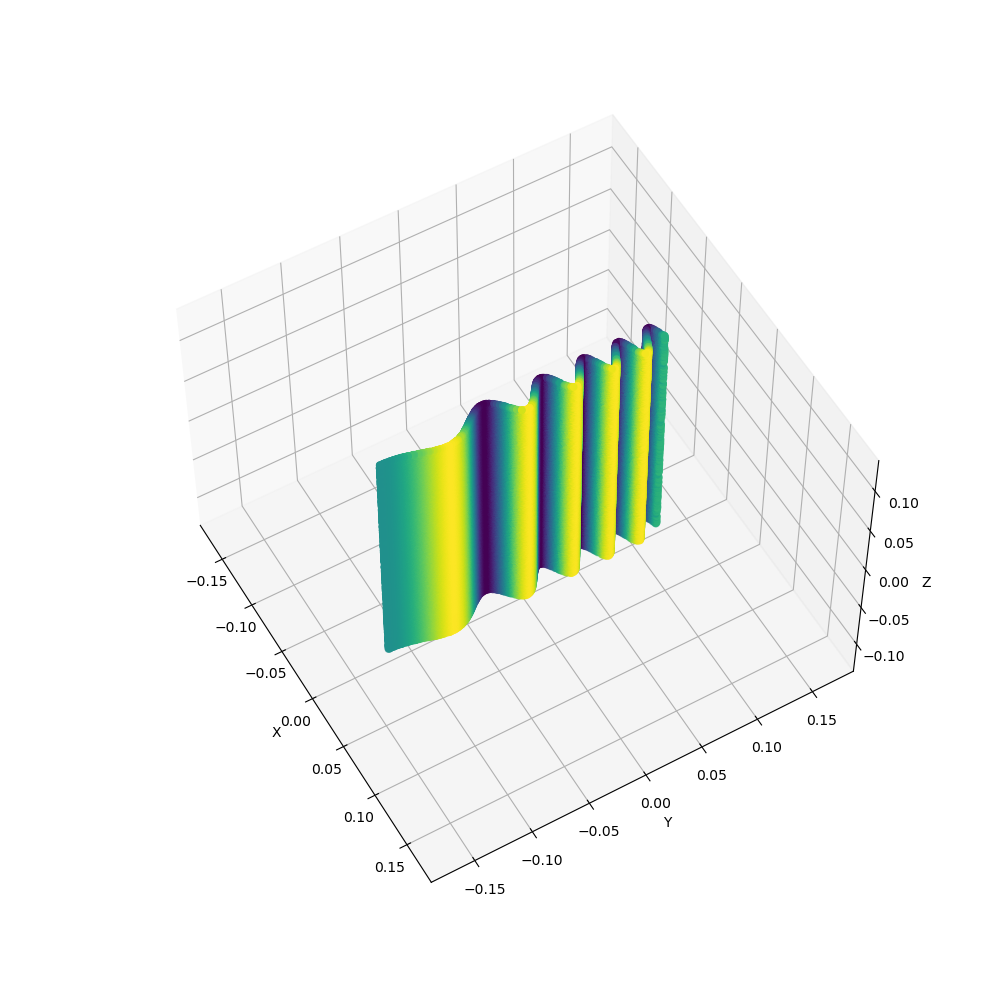}
    \caption{Ground-truth 3D sheet geometry}
    \label{fig:coarse_fine_sheet_3d}
  \end{minipage}
  \hspace{5pt}
  \begin{minipage}[b]{0.23\linewidth}
    \centering
    \includegraphics[width=\linewidth, angle=-90]{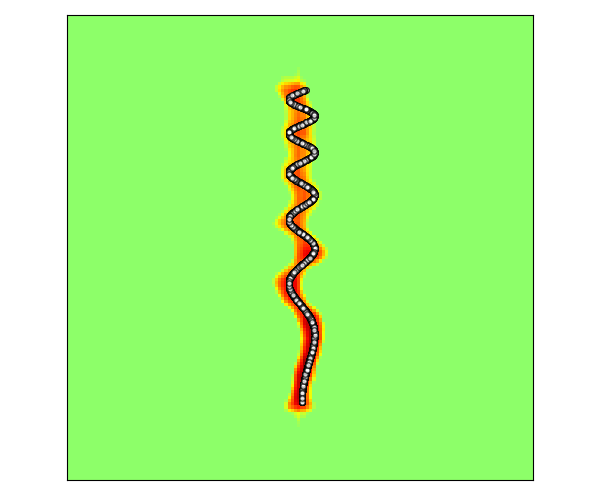}
    \caption{Overlay of predicted maxima (red) with GT (white)}
    \label{fig:coarse_fine_sheet_2d}
  \end{minipage}
  \hspace{5pt}
  \begin{minipage}[b]{0.45\linewidth}
    \centering
    \includegraphics[width=\linewidth]{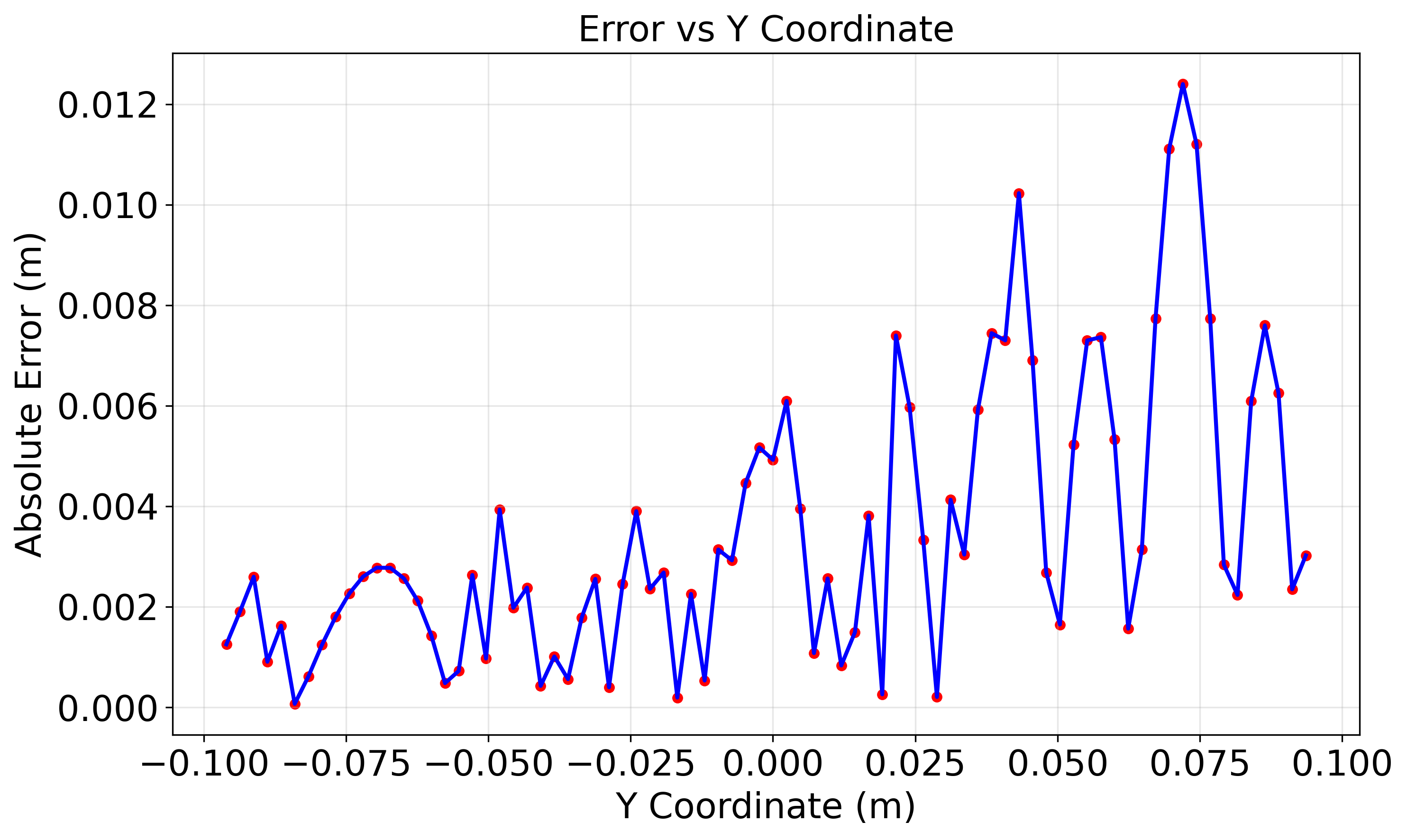}
    \caption{Error vs. vertical Y-axis}
    \label{fig:error_vs_y}
  \end{minipage}
  \label{fig:coarse_fine_experiment}
\end{figure}

To evaluate the spatial resolution limits of \systemname{}, we introduce a synthetic benchmark object designed to gradually vary in surface detail. Specifically, we construct a vertical sheet where sinusoidal undulations increase in frequency along the y-axis (Fig.~\ref{fig:coarse_fine_sheet_3d}). The initial part of the object features smooth, low-frequency ripples, while the latter part contains increasingly fine patterns that challenge the system's resolving capacity.

We render volumetric reconstructions using \systemname{} and compare the learned scatterer field with the ground-truth surface. As shown in Fig.~\ref{fig:coarse_fine_sheet_2d}, \systemname{} accurately captures coarse structures in the first half, with clear alignment between the predicted maxima and the ground-truth scatterer locations. However, in the high-frequency regions, the system begins to miss fine oscillations, resulting in a loss of structural fidelity and localized blurring.

To quantify this degradation, we extract the predicted maxima along each vertical slice and compute their Euclidean distance to the corresponding ground-truth scatterer points. Fig.~\ref{fig:error_vs_y} plots the absolute reconstruction error as a function of the vertical Y-axis coordinate. We observe that errors remain low (below 4 mm) in low-detail regions but increase steadily, exceeding 1 cm in the highest-frequency zones. This highlights a fundamental resolution ceiling tied to the system's physical model and the radar's bandwidth.

\subsubsection{Impact of Different Losses on Reconstruction}

To better understand the contribution of each loss component—magnitude, real, and imaginary—we analyze their sensitivity to spatial perturbations in the scatterer field. Specifically, we add increasing levels of Gaussian noise to the ground-truth scatterer locations and observe the corresponding change in normalized MSE across the three loss terms.

\begin{wrapfigure}{r}{0.4\textwidth}
    \vspace{-8pt}
    \includegraphics[width=\linewidth]{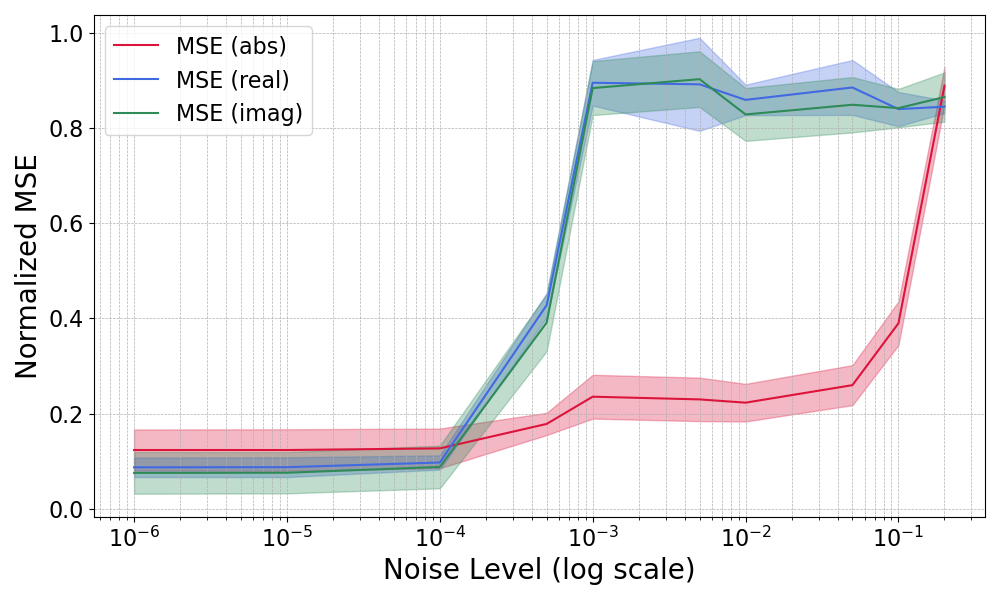}
    \caption{Effect of noise level on MSE metrics across different components.}
    \label{fig:different_losses}
    \vspace{-5pt}
\end{wrapfigure}

As shown in Figure~\ref{fig:different_losses}, the absolute magnitude loss (\textit{abs}) exhibits strong gradients even at large perturbation scales (on the order of 10 cm), making it well-suited for guiding the network toward a coarse geometry early in training. On the other hand, losses on the real and imaginary components become more discriminative only at finer perturbation scales (on the order of 1 mm), where they help capture high-frequency phase structure.

Motivated by this behavior, our training schedule applies the \textit{abs} loss during the initial iterations to encourage coarse alignment. Once the model has converged to a reasonable estimate, we introduce a combined loss that includes both the real and imaginary terms, enabling the network to refine fine-grained details. In practice, we apply the magnitude-only loss for the first 10\% of transmitter locations, followed by a weighted combination of all three components. This staged supervision strategy empirically improves convergence and reconstruction fidelity across a range of SNR conditions.

\section{Conclusion}
In this paper, we introduced \systemname{}{}, a novel framework for volumetric reconstruction using Frequency-Modulated Continuous-Wave (FMCW) radar data. By integrating a fully differentiable forward model operating in the frequency domain with implicit neural representations (INRs), \systemname~effectively leverages the inherent linear relationship between beat frequency and scatterer distance. This approach not only facilitates more efficient and accurate learning of scene geometry but also enhances computational efficiency by focusing on relevant frequency bins.

Our extensive experiments demonstrate that \systemname{} significantly outperforms classical backprojection methods and existing learning-based approaches, achieving higher resolution and more accurate reconstructions of complex scenes. This advancement represents the first application of neural volumetric reconstruction in the radar domain, offering a promising direction for future research in radar-based imaging and perception systems.

Future work could explore extending \systemname~to dynamic scenes, incorporating motion estimation to handle moving objects. Additionally, integrating \systemname~with other sensor modalities, such as LiDAR or optical cameras, could further enhance reconstruction accuracy and robustness. 

\section{Acknowledgement}
This work was partially supported by NSF CAREER Award 2238433. We also thank the various companies that sponsor the iCoSMoS laboratory at UMD.



\clearpage

\end{document}